\documentclass[journal]{IEEEtran}
\usepackage[inline]{trackchanges}

\usepackage[hyphens]{url}  
\usepackage{graphicx} 
\usepackage[square,numbers]{natbib}  
\usepackage{caption} 
\usepackage{algorithm}
\usepackage{algorithmic}
\usepackage{amsmath}
\usepackage{amsthm}
\newtheorem{definition}{Definition}
\usepackage{amssymb}
\usepackage{amsfonts}
\usepackage{bm}
\usepackage{array}
\usepackage{multirow}
\usepackage[english]{babel}
\usepackage{adjustbox}
\usepackage{subfigure}
\usepackage{booktabs}
\usepackage{enumitem}
\usepackage{newfloat}
\usepackage{listings}

\usepackage{epstopdf}
\usepackage{mathrsfs}

\definecolor{cvprblue}{rgb}{0.21,0.49,0.74}
\usepackage[pagebackref,breaklinks,colorlinks,allcolors=cvprblue]{hyperref}

\makeatletter
\g@addto@macro{\UrlBreaks}{\UrlOrds}
\makeatother
\usepackage[flushleft]{threeparttable} 

\newtheorem{prbl}{Problem}

\begin{document}

\title{Activity-aware Human Mobility Prediction with Hierarchical Graph Attention Recurrent Network}

\author{Yihong~Tang*,
        Junlin~He*,
        Zhan~Zhao$^\dagger$
\thanks{Y. Tang is with the Department of Urban Planning and Design, The University of Hong Kong, Hong Kong SAR, China (E-mail: yihongt@connect.hku.hk).}
\thanks{J. He is with the Department of Civil and Environmental Engineering, The Hong Kong Polytechnic University, Hong Kong SAR, China (E-mail: junlin.he@polyu.edu.hk).}
\thanks{Z. Zhao is with the Department of Urban Planning and Design, and the Musketeers Foundation Institute of Data Science, The University of Hong Kong, Hong Kong SAR, China (E-mail: zhanzhao@hku.hk)}
\thanks{$^\dagger$ Corresponding author. * Equal Contributions.}
}

\markboth{IEEE Transactions on Intelligent Transportation Systems, 2024}{}

\maketitle

\begin{abstract}
 
Human mobility prediction is a fundamental task essential for various applications in urban planning, location-based services and intelligent transportation systems.
Existing methods often ignore activity information crucial for reasoning human preferences and routines, or adopt a simplified representation of the dependencies between time, activities and locations. 
To address these issues, we present \textit{\textbf{H}ierarchical \textbf{G}raph \textbf{A}ttention \textbf{R}ecurrent \textbf{N}etwork} (\textbf{\textsc{Hgarn}}) for human mobility prediction. Specifically, we construct a hierarchical graph based on past mobility records and employ a \textit{Hierarchical Graph Attention Module} to capture complex time-activity-location dependencies. This way, \textsc{Hgarn} can learn representations with rich human travel semantics to model user preferences at the global level. We also propose a model-agnostic history-enhanced confidence (\textsc{MaHec}) label to incorporate each user’s individual-level preferences.
Finally, we introduce a \textit{Temporal Module}, which employs recurrent structures to jointly predict users’ next activities and their associated locations, with the former used as an auxiliary task to enhance the latter prediction.
For model evaluation, we test the performance of \textsc{Hgarn} against existing state-of-the-art methods in both the \textit{recurring} (i.e., returning to a previously visited location) and \textit{explorative} (i.e., visiting a new location) settings. 
Overall, \textsc{Hgarn} outperforms other baselines significantly in all settings based on two real-world human mobility data benchmarks. These findings confirm the important role that human activities play in determining mobility decisions, illustrating the need to develop activity-aware intelligent transportation systems.
Source codes of this study are available at \url{https://github.com/YihongT/HGARN}.
\end{abstract}


\IEEEpeerreviewmaketitle
\section{Introduction}\label{sec:intro}

Human mobility is critical for various downstream applications such as urban planning, location-based services and intelligent transportation systems. The ability to model and accurately predict future human mobility can inform important public policy decisions for managing traffic congestion, promoting social integration, and maximizing productivity \cite{schlapfer2021universal}. Central to human mobility modeling is the problem of next location prediction, i.e., predicting where an individual is going next, which has received great attention in research and practices. On the one hand, the increasing prevalence of mobile devices and popularity of location-based social networks (LBSNs) provide unprecedented data sources for mining individual-level mobility traces and preferences \cite{yang2014modeling}. On the other hand, the advancement of AI and machine learning offers a plethora of analytical tools for modeling human mobility. These innovations greatly enhance human mobility modeling in the past decade, especially for the next location prediction.

\begin{figure}[t]
    \centering
    \includegraphics[width=.49\textwidth]{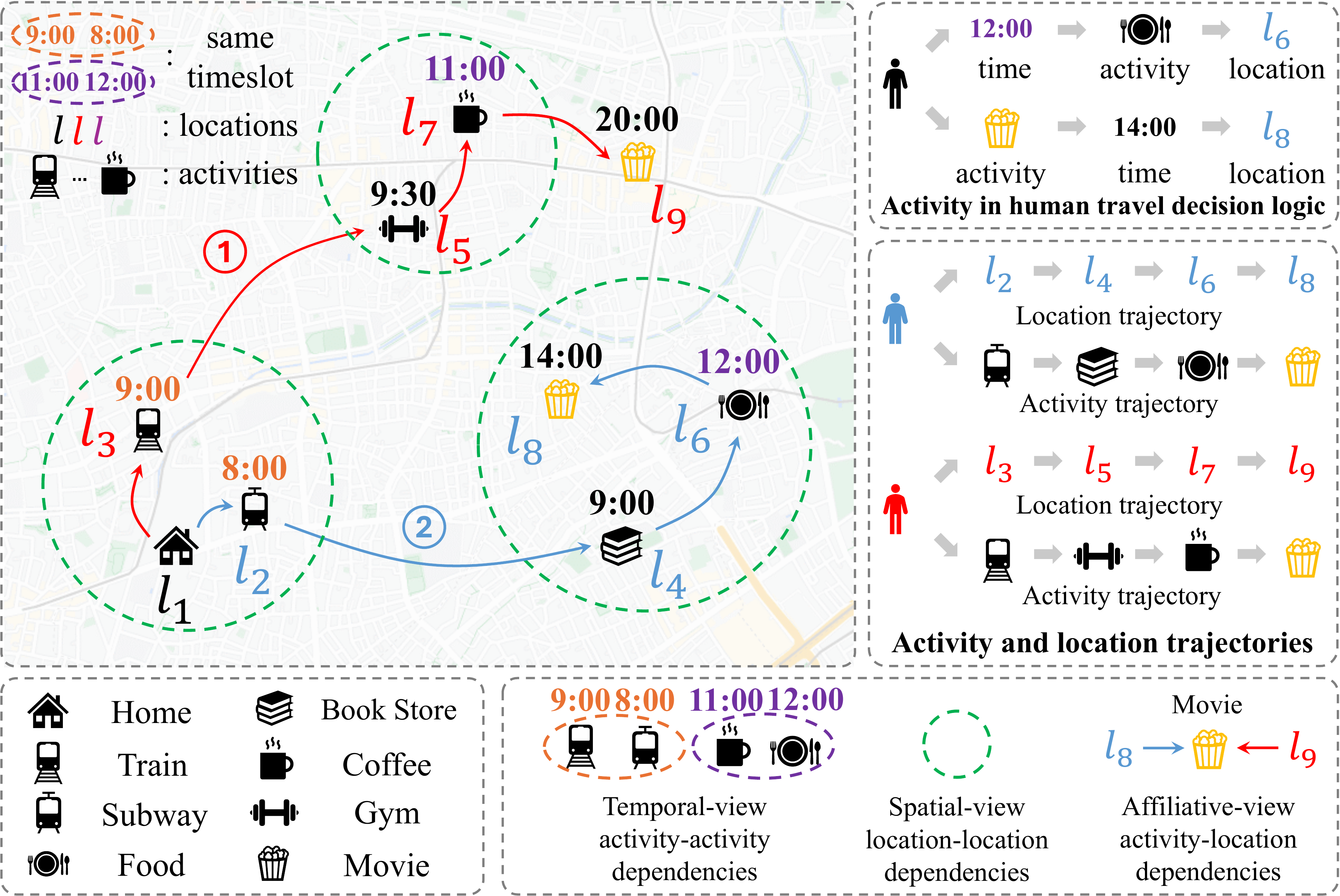}
    \caption{An illustration of two human mobility trajectories. Activities are essential in affecting human travel decisions.}
    \label{fig:illustration}
\end{figure}

While traditional approaches to human mobility analysis typically used Markov Chains (\textsc{Mc}) \cite{gambs2012next,rendle2010factorizing,mo2021individual} to model transition patterns over location sequences, recurrent neural networks (\textsc{Rnn}) \cite{hochreiter1997long} demonstrated superior predictive performance, including pioneering works that employed recurrent structures to model temporal periodicity \cite{feng2018deepmove} and spatial regularity \cite{yang2020location}. 
Due to the great success of the \textsc{Transformer} architecture \cite{vaswani2017attention}, the attention mechanism has also been adopted to model sequences and obtain competitive prediction results \cite{luo2021stan,guo2020attentional}. In recent years, graph-based approaches leveraged graph representation learning \cite{yang2019revisiting,chang2020learning} and graph neural networks (\textsc{Gnn}s) \cite{kipf2016semi} to model user preferences \cite{lim2020stp} and spatial-temporal relationships \cite{dang2022predicting} between locations, obtaining rich representations \cite{wang2021spatio,rao2022graph} to improve the performance of next location prediction.
However, most existing studies focus on predicting human mobility based on individual location sequences, overlooking the integral interplay between activity participation and location visitation behaviors. Classic travel behavior theories suggest that an individual's travel decisions are determined by the need to participate in activities taking place at different locations and scheduled at different time of day \cite{castiglione2015activity}. Given that human activity data is becoming increasingly accessible and most location visits can be characterized by only a small number of activity categories, incorporating these activity dynamics into human mobility modeling offers a behaviorally insightful and computationally efficient approach.

Figure \ref{fig:illustration} shows several human mobility trajectories reflecting time-activity-location dependencies. For example, when the time is approaching noon, one user may dine at a nearby restaurant, and another may go to the movie theater for a specific starting time, which illustrates that activities are usually scheduled according to the time of day. 
People typically make location decisions based on the intended activities, and thus considering activity information can lead to better predictability of human mobility.
However, few studies have considered activity information (e.g., location categories) for next location prediction. 
Notably, \textsc{Cslsl}, proposed by Huang et al. \cite{huang2022human}, adopts an \textsc{Rnn}-based structure \cite{chung2014empirical} to model human travel decision logic, where the time, activity, and location are predicted sequentially. However, the design of \textsc{Cslsl} oversimplifies the time-activity-location dependencies. 
Given data sparsity and behavioral uncertainties, the time prediction tends to be more challenging \cite{zhao2018individual}, 
which may compromise the prediction of activities and locations.

Based on the above observations, a suitable human next location predictor should: 
\textbf{(1) take into account human activities when predicting next locations}, leveraging the predicted future activity information to enhance location prediction, and (2) \textbf{effectively manage intricate time-activity-location dependencies} while circumventing the difficulty in time prediction under data sparsity and uncertain human behaviors.
In this study, we propose \textit{\textbf{H}ierarchical \textbf{G}raph \textbf{A}ttention \textbf{R}ecurrent \textbf{N}etwork} (\textbf{\textsc{Hgarn}}) for next location prediction. Specifically, we construct a hierarchical graph based on past mobility records and employ a \textit{Hierarchical Graph Attention Module} to capture complex time-activity-location dependencies. This way, \textsc{Hgarn} can learn representations with rich human travel semantics to model user preferences at the global level. We also propose a model-agnostic history-enhanced confidence (\textsc{MaHec}) label to incorporate each user’s individual-level preferences. Finally, we introduce a \textit{Temporal Module}, which employs recurrent structures to jointly predict users’ next activities and their associated locations, with the former used as an auxiliary task to enhance the latter prediction.
Through such design, \textsc{Hgarn} can leverage the learned time-activity-location dependencies to benefit both global- and individual-level human mobility modeling, and use predicted next activity distribution to facilitate next location prediction. 
In summary, this study makes the following contributions:

\begin{itemize}[leftmargin=*]
    \item We propose a Hierarchical Graph that incorporates human activity information to represent the activity-activity, activity-location, location-location dependencies. To our best knowledge, among the few methods considering activity information for next location prediction, this is the first work to model the dependencies of time, activities and locations using a Hierarchical Graph.
    \item We design a activity-aware Hierarchical Graph Attention Recurrent Network (\textsc{Hgarn}), which contains a \textit{hierarchical graph attention module} to model dependencies between time, activities, and locations, and a \textit{temporal module} to incorporate the hierarchical graph representations into sequence modeling, leveraging next activity prediction to boost next location prediction.
    \item We introduce a simple yet effective model-agnostic history-enhanced confidence (\textsc{MaHec}) label to guide the model learning of each user's individual-level preferences, allowing the model to focus more on relevant locations in their history trajectories when predicting their next locations.
    \item Through extensive experiments, we evaluate the prediction performance of \textsc{Hgarn} against existing SOTAs in both the \textit{recurring}, and \textit{explorative} settings, using two real-world LBSN check-in datasets. Our work is the first to separately evaluate next location prediction performance in these settings. The results show that \textsc{Hgarn} can significantly outperform all baselines in all experimental settings.
\end{itemize}

\begin{figure*}[t]
    \centering
    \includegraphics[width=0.95\textwidth]{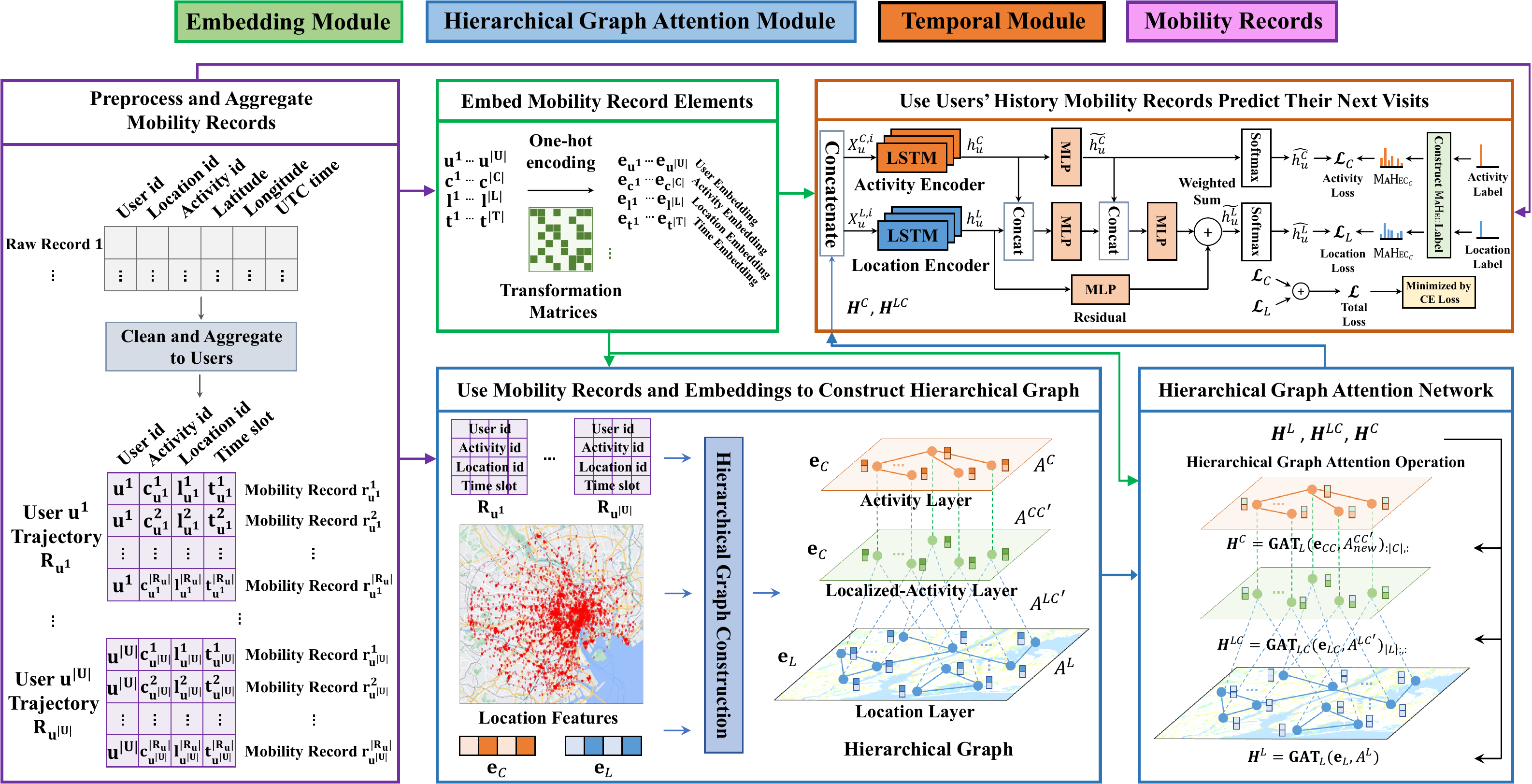}
    \caption{A workflow of the proposed \textsc{Hgarn}.}
    \label{fig:model}
\end{figure*}

\section{Related work}

\subsection{Next Location Prediction}

Next location prediction is essentially about sequence modeling since the next location visit is usually dependent on the previous one \cite{cheng2013you,zhang2014lore}. Traditional \textsc{Mc}-based methods often incorporate other techniques, such as matrix factorization \cite{rendle2010factorizing} and activity-based modeling \cite{mo2021individual}, for enhanced prediction performance. However, they are limited in capturing long-term dependencies or predicting \textit{explorative} human mobility.

\textsc{Rnn}-based models regard the next location prediction problem as a sequence-to-sequence task and have shown superior performance. \textsc{Strnn} \cite{liu2016predicting} is a pioneering work that integrates spatial-temporal characteristics between consecutive human visits into \textsc{Rnns}, laying the groundwork for subsequent studies. Building on this, \textsc{Stgn} \cite{zhao2020go} introduces spatial and temporal gates to Long Short-Term Memory (\textsc{Lstm}) networks to better capture users' interests, while \textsc{Flashback} \cite{yang2020location} leverages spatial and temporal intervals to aggregate past RNN hidden states for improved predictions. Additionally, \textsc{Lstpm} \cite{sun2020go} employs a non-local network and a geo-dilated \textsc{Lstm} to model both long- and short-term user preferences. Attention mechanisms are also utilized to enhance model performance. \textsc{DeepMove} \cite{feng2018deepmove} combines attention mechanisms with \textsc{Rnn} modules to effectively capture users' long- and short-term preferences. Similarly, \textsc{Arnn} \cite{guo2020attentional} uses a knowledge graph to identify related neighboring locations and employs attentional \textsc{Rnns} to model the sequential regularity of check-ins. Furthermore, \textsc{Stan} \cite{luo2021stan} extracts relative spatial-temporal information between both consecutive and non-consecutive locations through a spatio-temporal attention network. These approaches collectively highlight the importance of integrating spatial-temporal dynamics and attention mechanisms to improve the accuracy of human mobility predictions. In addition, some efforts incorporate contextual information \cite{li2018next} such as geographical information \cite{lian2020geography}, dynamic-static \cite{manotumruksa2017deep}, text content about locations \cite{chang2018content} into sequence modeling.

Graph-based models have become a cornerstone in the field of human mobility prediction due to their ability to effectively capture complex relationships and dependencies. For instance, \textsc{LBSN2Vec} \cite{yang2019revisiting} employs random walks on a hypergraph to learn embeddings, enhancing predictions for both locations and friendships. Similarly, \textsc{STP-UDGAT} \cite{lim2020stp} leverages graph attention networks (\textsc{Gat}) to discern location relationships from both local and global perspectives, utilizing spatial, temporal, and preference graphs. To address the data sparsity issue, \textsc{Hmt-Grn} \cite{lim2022hierarchical} constructs multiple user-region matrices at varying granularities to improve prediction accuracy. \textsc{Gcdan} \cite{dang2022predicting} integrates graph convolutional networks (\textsc{Gcn}) to capture high-order sequential dependencies in a dual attention framework to mitigate sparsity. \textsc{Graph-Flashback} \cite{rao2022graph} innovatively combines knowledge graph embeddings with \textsc{Gcn} to refine graph representations, further integrating with the \textsc{Flashback} model for enhanced prediction capabilities. Moreover, to provide more activity awareness in human mobility modeling, generative adversarial imitation learning (\textsc{Gail}) has been adapted to simulate activity trajectories \cite{yuan2022activity}. Other significant contributions include the use of weighted category hierarchy in \cite{bao2012location} to model activities, \textsc{CatDM} \cite{yu2020category} which incorporates activities and spatial distances to reduce search space, and \textsc{Cslsl} \cite{huang2022human} that introduces an \textsc{Rnn}-based causal structure to capture the logic behind human travel decisions. However, most existing methods overlook the activity information and cannot effectively model the time-activity-location dependencies, which are essential for predicting and understanding human mobility.

\subsection{Hierarchical Graph Neural Network}
The Hierarchical Graph Neural Network (\textsc{HGnn}) is a family of \textsc{Gnn} models that gain significant attention in recent years due to their ability to capture complex dependencies in data using hierarchical structures. \textsc{HGnn} has been applied to various urban applications such as parking availability prediction \cite{zhang2020semi}, air quality forecasting \cite{xu2021highair}, road network representation learning \cite{wu2020learning}, real estate appraisal \cite{zhang2021mugrep}, and socioeconomic indicator estimation \cite{zhou2023hierarchical}. However, each model has its own structure design and graph construction mechanisms based on their specific application scenarios, resulting in fundamentally different architectures.

One relevant \textsc{HGnn}-based approach for next location prediction is \textsc{Hmt-Grn} \cite{lim2022hierarchical}, which partitions the spatial map and performs a Hierarchical Beam Search (\textsc{Hbs}) on different regions and POI distributions to hierarchically reduce the search space for accurate predictions. Unlike previous works, our proposed \textsc{Hgarn} is an activity-based model designed for next location prediction. It constructs a hierarchical graph based on human activities and leverages graph attention mechanisms to capture complex time-activity-location dependencies. This activity-based design is unique and distinguishes \textsc{Hgarn} from other \textsc{HGnn}-based models.

\section{Preliminaries}

We use notations $U=\left \{ u^{i} \right \}_{i=1}^{|U|}$, $L=\{l^{i}\}_{i=1}^{|L|}$, $C=\{c^{i}\}_{i=1}^{|C|}$, and $T=\{t^{i}\}_{i=1}^{|T|}$ to denote the sets of users, locations, activities and time series, respectively. For a specific user $u\in U$, we denote their sets of locations, activities and time series in a temporal order as $L_{u}=\{l^{i}_{u} \}_{i=1}^{|L_{u}|}$, $C_{u}=\{c^{i}_{u} \}_{i=1}^{|C_{u}|}$, and $T_{u}=\{t^{i}_{u} \}_{i=1}^{|T_{u}|}$, respectively. 

\begin{definition}[\textbf{Mobility Record}]
Let us use $r$ to denote a single human mobility record. Each mobility record comprises a user $u \in U$, an activity $c_{u}^{i} \in C_{u}$, a location $l_{u}^{i} \in L_{u}$ and the visit time $t_{u}^{i} \in T_{u}$. The $i$th record of user $u$ is thus represented by a tuple $r_{u}^{i}=(u, c_{u}^{i}, l_{u}^{i}, t_{u}^{i})$. 
\end{definition}

\begin{definition}[\textbf{Trajectory}]

A trajectory is a sequence of mobility records for a user $u$, denoted by $R_{u}=\{r_{u}^{i}\}_{i=1}^{|R_{u}|}$. Each trajectory $R_{u}$ can be divided into a activity trajectory $R_{u}^{C}={\{c_{u}^{i}\}}_{i=1}^{|R_{u}|}$, location trajectory $R_{u}^{L}={\{l_{u}^{i}\}}_{i=1}^{|R^{u}|}$, and time trajectory $R_{u}^{T}={\{t_{u}^{i}\}}_{i=1}^{|R_{u}|}$.
\end{definition}

\begin{prbl}[\textbf{Next Location Prediction}]
Given a user $u$'s observed trajectory $R_{u}$ as input, we consider $u$'s next record as its future state. The human mobility prediction task $\mathcal{T}$ maps $u$'s past trajectory $R_{u}$ to $u$'s next location $l_{u}^{|R_{u}|+1}$ in the future. The problem can be expressed as follows:
\begin{equation} \label{eq1}
    R_{u}\stackrel{\mathcal{T}(\cdot;\theta)}{\longrightarrow}l_{u}^{|R_{u}|+1},
\end{equation}
where $\theta$ is the parameters of mapping $\mathcal{T}$.
\end{prbl}

\section{Methodology}

\textsc{Hgarn}'s workflow is demonstrated in Figure \ref{fig:model}. The raw data is first encoded in the \textit{embedding module} and then input to the \textit{hierarchical graph attention module} to model multi-dependencies. Finally, the user's personalized embeddings are fused with the learned hierarchical graph representations and input to the \textit{temporal module} to make predictions.
We will elaborate on the details of \textsc{Hgarn} in the following sections.

\subsection{Embedding Module}

The \textit{embedding module} aims to learn low-dimensional embedding vectors to represent each user, activity, location and time interval. It is worth noting that the first three elements are all naturally discrete, and the continuous time can be discretized into time intervals as well, making it easier to learn embedding vectors. 
In this work, the time is represented by two discrete variables, one for the hour of day $h \in T^{h}$ and the other for the day of week $w \in T^{w}$. Note that all $t \in T$ can be written in the form of $t=(h, w)$.

To illustrate how we generate the trainable embedding vectors used for next location prediction, we first represent users, activities, locations, and time intervals as one-hot encoded vectors. Specifically, we define the one-hot vectors as follows: $\boldsymbol{v}_u \in \mathbb{R}^{1 \times |U|}$ for users, $\boldsymbol{v}_l \in \mathbb{R}^{1 \times |L|}$ for locations, $\boldsymbol{v}_c \in \mathbb{R}^{1 \times |C|}$ for activities, and $\boldsymbol{v}_t \in \mathbb{R}^{1 \times |T|}$ for time intervals. In each one-hot vector, only one element is set to 1, with all other elements being 0. This single 1 uniquely identifies the corresponding entity (e.g., a specific user or location).
To convert these high-dimensional discrete one-hot vectors into low-dimensional continuous trainable embeddings for actual use, we apply the following transformations:
\begin{equation} \label{eq2}
    \boldsymbol{e}_u = \boldsymbol{v}_u \boldsymbol{W}_u; \boldsymbol{e}_l = \boldsymbol{v}_l \boldsymbol{W}_l; \boldsymbol{e}_c = \boldsymbol{v}_c \boldsymbol{W}_c; \boldsymbol{e}_t = \boldsymbol{v}_t \boldsymbol{W}_t, 
\end{equation}
where $\boldsymbol{e}_u \in \mathbb{R}^{1 \times d^u}$, $\boldsymbol{e}_l \in \mathbb{R}^{1 \times d}$, $\boldsymbol{e}_c \in \mathbb{R}^{1 \times d}$, and $\boldsymbol{e}_t \in \mathbb{R}^{1 \times d^t}$ represent the resulting embedding vectors for users, locations, activities, and time intervals, respectively. These embeddings are trainable and allow us to effectively capture latent information about each entity.
$\boldsymbol{W}_u \in \mathbb{R}^{|U| \times d^u}$, $\boldsymbol{W}_l \in \mathbb{R}^{|L| \times d}$, $\boldsymbol{W}_c \in \mathbb{R}^{|C| \times d}$, and $\boldsymbol{W}_t \in \mathbb{R}^{|T| \times d^t}$ are the corresponding transformation matrices, which can be learned jointly with other model parameters through back propagation. 
$d^u$, $d$, and $d^t$ are hyperparameters that denote the dimensions of the embedding vectors for users, activities/locations, and time intervals, respectively. After the transformation, the resulting embedding vectors $\boldsymbol{e}_u, \boldsymbol{e}_l, \boldsymbol{e}_c$, and $\boldsymbol{e}_t$ are typically ``squeezed" to remove the extra dimension, resulting in vectors $\boldsymbol{e}_u \in \mathbb{R}^{d^u}$, $\boldsymbol{e}_l \in \mathbb{R}^{d}$, $\boldsymbol{e}_c \in \mathbb{R}^{d}$, and $\boldsymbol{e}_t \in \mathbb{R}^{d^t}$, respectively.
To illustrate the embedding process, we provide an example to learn user embeddings in Appendix \ref{appx:embedding}.

\subsection{Hierarchical Graph Attention Module}

To model the complex dependencies between activities and locations, the \textit{hierarchical graph attention module} is designed with two parts: hierarchical graph construction and hierarchical graph attention networks for multi-dependencies modeling.

\subsubsection{Hierarchical Graph Construction}

The urban spatial network can be represented as a graph. \textsc{Gnn}s provide an effective way to learn graph representation and model node-to-node dependencies \cite{lv2021we,velivckovic2017graph}. In this study, we model the location-location, location-activity, and activity-activity dependencies using a hierarchical graph, 
which consists of three layers: the location layer, localized-activity layer, and activity layer. Here, the localized-activity layer is used to suppress noise aggregated from the location layer.

We formally describe the hierarchical graph with notation $\mathcal{G}=(\mathcal{V},\mathcal{E})$, where $\mathcal{V}$ $=$ $\mathcal{V}^{L} \cup \mathcal{V}^{C} \cup \mathcal{V}^{C^\prime}$ and $\mathcal{E}=\{A^{L}$,$A^{C}$,$A^{LC^\prime}$,$A^{CC^\prime}\}$. Specifically, $\mathcal{V}^{L}$ and $\mathcal{V}^{C}$ represent the sets of location nodes and activity nodes, respectively. $\mathcal{V}^{C^\prime}$ indicates the set of localized-activity nodes, which is an identical copy of the activity node set. $\mathcal{E}$ comprises four adjacency matrices denoting the dependencies between two location nodes ($A^{L}$), two activity nodes ($A^{C}$), a location node and a localized-activity node ($A^{LC^\prime}$), and an activity node and a localized-activity node ($A^{CC^\prime}$).

The location adjacency matrix $A^{L}$ is defined based on the geographical distance between locations. Specifically, two locations $l^{i}$ and $l^{j}$ are linked with an edge if their \textit{haversine} distance is within a threshold. $A^{L} \in \mathbb{R}^{|L| \times |L|}$ is defined as:
\begin{equation}\label{eq3}
    A^{L}_{l^{i}, l^{j}}=
    \begin{cases}
        1, & Haversine \left(l^{i}, l^{j} \right) < D^{h}\\
        0, & \text{otherwise}
    \end{cases},
\end{equation}
where $D^{h}$ is a hyperparameter denoting the distance threshold.

The construction of $A^{C}$ is based on observed trajectories. Intuitively, the dependencies between activities can be measured by the frequency of co-occurrence in the same time interval. However, if we directly consider the activity co-occurrence frequency based on all trajectories from all users, it may lead to unrelated activities being linked (e.g., check-in at subway stations and gyms both often occur in the evening), due to the difference in user preferences. Instead, we propose to learn the inter-activity dependencies based on activity co-occurrence within individual-level trajectory sets. Therefore, we can traverse each user's trajectories $R_{u}$ and count the co-occurrence frequency $M^{C}_{c^{i}, c^{j}}$ between each activity pair $(c^{i}, c^{j})$.
Based on activity co-occurrence frequencies, $A^{C} \in \mathbb{R}^{|C| \times |C|}$ is defined as:
\begin{equation}\label{eq4}
    A^{C}_{c^{i}, c^{j}}=
    \begin{cases}
        1, & \text{if } M^{C}_{c^{i}, c^{j}} > \text{mean}\left(M^{C}\right) \\
        0, & \text{otherwise}
    \end{cases},
\end{equation}

The adjacency matrix $A^{LC^{\prime}}$ defines the dependencies between location nodes and localized-activity nodes. Each node of $V^{L}$ is linked to only one node of $V^{C^\prime}$, representing the corresponding activity category at that location. In contrast, each node of $V^{C^\prime}$ may be linked to multiple nodes of $V^{L}$, as several locations can share the same activity type. Formally, we define $A^{LC^{\prime}}_{L} \in \mathbb{R}^{|L| \times |C|}$ based on the affiliations of locations and activities, where each row corresponds to a location and each column an activity. Additionally, 
we construct the adjacency matrix $A^{LC^{\prime}} \in \mathbb{R}^{(|L|+|C|) \times (|L|+|C|)}$ based on $A^{LC^{\prime}}_{L}$ in the following block matrix form:
\begin{equation}\label{eq5}
A^{LC^{\prime}}=\begin{bmatrix}
O_{L} & A^{LC^{\prime}}_{L} \\
{(A^{LC^{\prime}}_{L})}^{\top} & O_{C}
\end{bmatrix},
\end{equation}
where ${(A^{LC^{\prime}})}^{\top}$ is the transpose of $A^{LC^{\prime}}$, $O_{L} \in \mathbb{R}^{|L| \times |L|}$ and $O_{C} \in \mathbb{R}^{|C| \times |C|}$ are two zero matrices.

The localized-activity layer is designed to suppress noise from the location layer aggregated to the activity layer. Therefore, each node in the localized-activity layer is assumed to be connected to the node in the activity layer representing the same activity type. Similarly, we have $A^{CC^\prime} \in \mathbb{R}^{2|C| \times 2|C|}$:
\begin{equation}\label{eq6}
    A^{CC^\prime}=\begin{bmatrix}
        O_{C} & I_{C} \\
        I_{C} & O_{C}
    \end{bmatrix},
\end{equation}
where $I_{C} \in \mathbb{R}^{|C| \times |C|}$ is an identity matrix.

\subsubsection{Hierarchical Graph Attention Networks}

\textsc{Gnn}s have proven to be powerful in capturing dependencies on graphs. Both inter- and intra-layer nodes on the hierarchical graph have different dependencies on each other.
Since the importance of locations within a certain distance are different to each other, we use \textsc{Gat} to model location-location dependencies:
\begin{equation} \label{eq7}
    H^{L}=\textsc{\textsc{Gat}}_{L}\left(e_{L}, A^{L}\right),
\end{equation}
where $\textsc{\textsc{Gat}}(\cdot)$ is an implementation of the original model \cite{velivckovic2017graph}, $H^{L} \in \mathbb{R}^{|L| \times d^{g}}$ is the learned representations as the output of the $\textsc{Gat}_{L}$.

To integrate location information into representation learning of activities and suppress the noise aggregated to the nodes of activity layer, we introduce the localized-activity layer to pre-aggregate location embeddings. We first concatenate $e_L$ and $e_C$ to obtain the fused embedding matrix $e_{LC} \in \mathbb{R}^{(|L|+|C|) \times d} = {\begin{bmatrix}e_L & e_C\end{bmatrix}}^{\top}$. Then the localized-activity process is implemented as: 
\begin{equation} \label{eq8}
    H^{LC^{\prime}}=\textsc{Gat}_{LC}\left(e_{LC}, A^{LC^{\prime}}\right),
\end{equation}
where $H^{LC^{\prime}} \in \mathbb{R}^{(|L|+|C|) \times d^{g}}$ is the output of $\textsc{Gat}_{LC}$. To obtain the pre-aggregated representation matrix ${H^{LC}} \in \mathbb{R}^{|C| \times d^{g}}$, we remove the first $|L|$ rows from $H^{LC^{\prime}}$.

The learned representation $H^{C^{\prime}}$ is again concatenated with activity embeddings $e_C$ as the $\textsc{Gat}_C$'s input $e_{CC}={\begin{bmatrix}e_C & H^{LC}\end{bmatrix}}^{\top}$. It is worth noting that for all nodes in the activity layer, we can simultaneously aggregate information from neighbors in the localized-activity layer and neighbors in the same layer by simply modifying the matrix $A^{CC^\prime}$ to:
\begin{equation} \label{eq9}
    A^{CC^\prime}_{new}=\begin{bmatrix}
        A^{C} & I_{C} \\
        I_{C} & O_{C}
    \end{bmatrix}.
\end{equation}

We employ a similar strategy to update the representation of the nodes in the activity layer:
\begin{equation} \label{eq10}
    {H^{C}}^\prime=\textsc{Gat}_{C}\left(e_{CC}, A^{CC^\prime}_{new}\right),
\end{equation}
where ${H^{C}}^\prime \in \mathbb{R}^{2|C| \times d^{g}}$ is the learned representation from $\textsc{Gat}_{C}$. To obtain the updated activity node representation $H^{C} \in \mathbb{R}^{|C| \times d^{g}}$, we remove the last $|C|$ rows from ${H^{C}}^\prime$.

The learned attention weights reflect the relative importance of each node to its neighbors, thereby demonstrating their influence within the network, and we provided more discussions in Section~\ref{res:ablation}.

\subsection{Temporal Module}

To model sequential dependencies of human mobility, the \textit{temporal module} is designed to encode a user's trajectory embeddings (from the embedding module) with learned graph representations (from the hierarchical graph attention module) through a recurrent structure. 
Given a user $u$'s trajectory, the learned representation for the $i$th activity or location can be denoted as:
\begin{equation} \label{eq11}
    \boldsymbol{X}^{C,i}_{u}= {\boldsymbol{e}_{u}} \Vert \boldsymbol{e}_{t_{u}^{i}} \Vert \boldsymbol{e}_{c_{u}^{i}} \Vert \boldsymbol{H}^{C}_{c_{u}^{i}},
\end{equation}
\begin{equation}\label{eq12}
    \boldsymbol{X}^{L,i}_{u}= {\boldsymbol{e}_{u}} \Vert \boldsymbol{e}_{t_{u}^{i}} \Vert \boldsymbol{e}_{l_{u}^{i}} \Vert \boldsymbol{H}^{C}_{c_{u}^{i}} \Vert \boldsymbol{H}^{L}_{l_{u}^{i}},
\end{equation}
where $\Vert$ is the concatenation operation, $\boldsymbol{H}^{C}_{c_{u}^{i}}$ and $\boldsymbol{H}^{L}_{l_{u}^{i}}$ are the learned graph representations of the activity node $c_{u}^{i}$ and location node $l_{u}^{i}$, respectively.

Specifically, \textsc{Lstm} is used to encode both user activity trajectories and location trajectories. Therefore, the hidden state updating process at $i$th iteration is implemented as:
\begin{equation} \label{eq13}
    \boldsymbol{c}^{C, i}_{u}, \boldsymbol{h}^{C, i}_{u} = \textsc{Lstm}(\boldsymbol{X}^{C,i}_{u},\boldsymbol{c}^{C, i-1}_{u}, \boldsymbol{h}^{C,i-1}_{u}),
\end{equation}
\begin{equation} \label{eq14}
    \boldsymbol{c}^{L, i}_{u}, \boldsymbol{h}^{L, i}_{u} = \textsc{Lstm}(\boldsymbol{X}^{L,i}_{u},\boldsymbol{c}^{L, i-1}_{u}, \boldsymbol{h}^{L,i-1}_{u}),
\end{equation}
where $\boldsymbol{h}^{C, i}_{u}$ and $\boldsymbol{h}^{L, i}_{u}$ are the $i$th hidden states for user $u$'s activity and location sequences, respectively. $\boldsymbol{c}^{C, i}_{u}$ and $\boldsymbol{c}^{L, i}_{u}$ are the corresponding cell states.

After obtaining final hidden states of activity and location encoder as $\boldsymbol{h}^{C}_{u}$ and $\boldsymbol{h}^{L}_{u}$, we implement our activity \textit{decoder} as a multi-layer perceptron (\textsc{Mlp}) to get the next activity logits $\widetilde{\boldsymbol{h}^{C}_{u}} \in \mathbb{R}^{|C|}$:
\begin{equation}\label{eq15}
    \widetilde{\boldsymbol{h}^{C}_{u}}=\textsc{Mlp}_{C}\left(\boldsymbol{h}^{C}_{u}\right),
\end{equation}
where the logits usually refer to the raw, unnormalized vectors output by the model, which are used as inputs to a Softmax function to obtain a predicted probability distribution.

Finally, we combine the obtained activity logits with the encoded representation $h^{L}_{u}$ using a residual connection \cite{he2016deep}. This results in the final location logits $\widetilde{h^{L}_{u}} \in \mathbb{R}^{|L|}$:
\begin{equation}\label{eq16}
    \begin{aligned}
    \widetilde{\boldsymbol{h}^{L}_{u}}= \lambda_{r}\cdot  & \textsc{Mlp}_{L}^{r}\left(\boldsymbol{h}^{L}_{u}\right)+ \\  (1-\lambda_{r})\cdot & \textsc{Mlp}_{L}\left(\textsc{Mlp}_{L}^{h}\left(\boldsymbol{h}^{L}_{u} \Vert \boldsymbol{h}^{C}_{u}\right) \Vert \widetilde{\boldsymbol{h}^{C}_{u}}\right),
    \end{aligned}
\end{equation}
where $\lambda_{r}$ is a factor that trades off different features.

\subsection{Model-Agnostic History-Enhanced Confidence Label}

Existing models \cite{feng2018deepmove,dang2022predicting} often try to learn temporal and periodic mobility patterns from collective trajectories from all users, overlooking the heterogeneity in individual preferences. Travel behavior theories suggest that individuals are more likely to revisit locations they have visited before, due to familiarity and established activity patterns \cite{sonmez1998determining}. 
To address this issue, we introduce a modified soft labeling approach known as the \textit{model-agnostic history-enhanced confidence} (\textsc{MaHec}) label. Unlike traditional labels which provide hard classifications, soft labels offer a probability distribution over classes, capturing uncertainty and allowing the model to learn more nuanced patterns \cite{hinton2015distilling}. The \textsc{MaHec} label incorporates historical user trajectory information, enhancing the model's ability to focus on relevant trajectories by assigning higher confidence to visited locations.

Specifically, for each location $l^{i}\in L$, we differentiate its confidence for a user $u$'s next location as follows:
\begin{equation}\label{eq17}
    {\textsc{MaHec}}_{l^{i}}^{u}=\begin{cases}
        w^{c}, & \text{if } l^{i}=l_{u}^{|R_{u}|+1} \\
        \left(1-w^{c}\right) \frac{f_{l^{i}}^{u}}{|R_{u}|}, & \text{if } l^{i} \in R_{u}^{L} \text{ and } l^{i} \neq l_{u}^{|R_{u}|+1}\\
        0, & \text{otherwise} 
    \end{cases}, 
\end{equation}
where $w^{c}\in[0,1]$ is a hyperparameter that indicates the confidence of $u$'s ground truth label, and $f_{l^{i}}^{u}$ denotes $u$'s frequency of visits to $l^{i}$ in the observed trajectory $R_{u}$.
Then the \textsc{MaHec} label for $u$'s next location is defined as:
\begin{equation}\label{eq18}
    {\textsc{MaHec}}_{L}^{u}=\left({\textsc{MaHec}}_{l^{i}}^{u},\right)_{i=1}^{|L|}\in \mathbb{R}^{|L|},
\end{equation}
where each element in ${\textsc{MaHec}}_{L}^{u}$ represents the confidence that the user $u$ decides to choose as their next location base on their past visits. Similarly, we conduct the same operations for user activity trajectories to obtain ${\textsc{MaHec}}_{C}^{u}$.

\subsection{Model Optimization}

Since next location prediction is a classification problem, we transform $\widetilde{h^{L}_{u}}$ from Eq.~\eqref{eq16} to the probability distribution of locations $\widehat{\boldsymbol{h}^{L}_{u}} \in \mathbb{R}^{|L|}$ through $\widehat{\boldsymbol{h}^{L}_{u}}=\text{Softmax}\left(\widetilde{\boldsymbol{h}^{L}_{u}}\right)$. 
Given ${\textsc{MaHec}}_{L}^{u}$ and $\widehat{\boldsymbol{h}^{L}_{u}}$, we can compute the cross-entropy loss for next location prediction, denoted as $\mathcal{L}_L$:
\begin{equation}\label{eq19}
    \mathcal{L}_L=- \frac{1}{|U|}\sum_{u \in U} \sum_{i=1}^{|L|}{\textsc{MaHec}}^{u}_{l^{i}}\cdot\log\left(\widehat{\boldsymbol{h}^{L, i}_{u}}\right),
\end{equation}
where $\widehat{\boldsymbol{h}^{L, i}_{u}}$ is the $i$th element of $\widehat{\boldsymbol{h}^{L}_{u}}$. Similarly, we compute the next activity loss $\mathcal{L}_{C}$ based on the same operations. Finally, we can train our \textsc{Hgarn} with a overall loss function:
\begin{equation}\label{eq20}
    \mathcal{L}=\lambda_{L}\cdot \mathcal{L}_{L}+\lambda_{C}\cdot \mathcal{L}_{C}
\end{equation}
where $\lambda_{L}$ and $\lambda_{C}$ are hyperparameters that trade off different loss terms.

\section{Experiments}

In this section, we compare \textsc{Hgarn} with existing SOTAs on two real-world LBSN check-in datasets.

\subsection{Datasets}

\renewcommand{\arraystretch}{1.2}
\begin{table}[h]
\caption{Statistical information of NYC and TKY datasets.} 
\centering 

\resizebox{0.48\textwidth}{!}{
\begin{tabular}{c|ccccc} \hline 
  & user & activity & location & trajectory & ratio (Rec / Exp) \\
\hline
NYC & 1065 & 308 & 4635 & 18918 & 85.9\% / 14.1\% \\
TKY & 2280 & 286 & 7204 & 49039 & 91.5\% / 8.5\% \\
\hline 
\end{tabular}          
}
\label{table:data_summary}
\end{table}

\begin{figure}[h]
    \centering
    \includegraphics[width=0.48\textwidth]{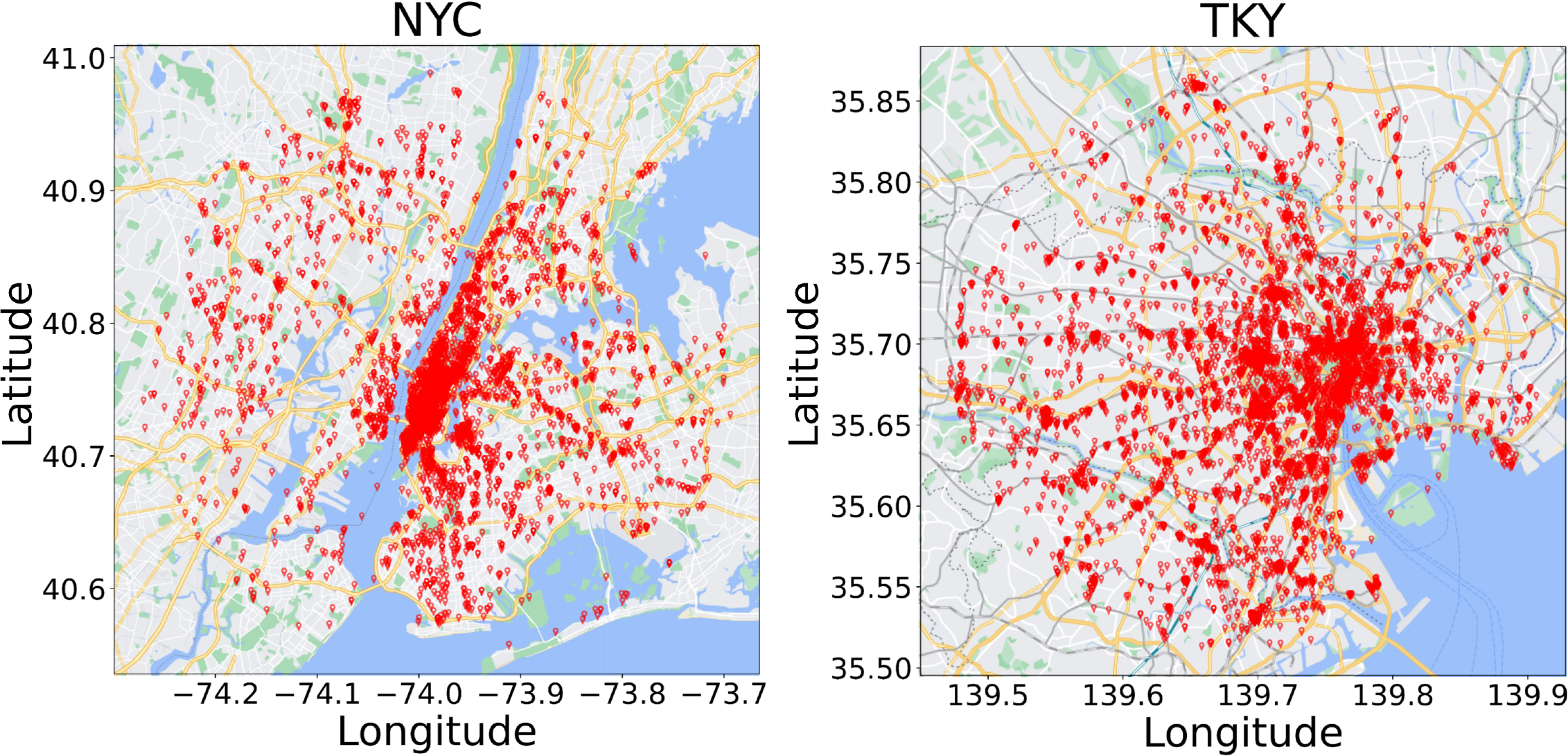}
    \caption{Location distributions of NYC and TKY.}
    \label{fig:spatial_appendix}
\end{figure}

We adopt two LBSN datasets \cite{yang2014modeling} containing Foursquare check-in records in New York City (NYC) and Tokyo (TKY) from April 12, 2012 to February 16, 2013, including 227,428 check-ins for NYC and 573,703 check-ins for TKY. The location distributions of NYC and TKY datasets are shown in Figure \ref{fig:spatial_appendix}. Users and locations with less than 10 records are removed following previous works. After cleaning, we obtain 308 and 286 activities for NYC and TKY, respectively. 
We divide the data into training and testing sets in a ratio of 8:2, following a chronological order (training first), in line with the conventions used in \cite{feng2018deepmove, zhao2020go, huang2022human}. Key data summary statistics are listed in Table~\ref{table:data_summary}.

\renewcommand{\arraystretch}{1.2}
\begin{table*}[t]
\caption{Main results. NYC \& TKY have activity info, different from Foursquare datasets used in some works, there reported results may not directly comparable to this work's. All experiments here report the best results within a consistent environment.} 
\centering 
\label{table2}
\resizebox{.95\textwidth}{!}{

\begin{threeparttable}
\begin{tabular}{c|ccc|ccc|ccc|ccc}
\toprule
\multirow{2}{*}{Main} & \multicolumn{6}{c|}{NYC}                                                           & \multicolumn{6}{c}{TKY}                                                         \\  \cline{2-13} 
                       & \multicolumn{1}{c}{R@1} & \multicolumn{1}{c}{R@5} & \multicolumn{1}{c|}{R@10} & \multicolumn{1}{c}{N@1} & \multicolumn{1}{c}{N@5} & \multicolumn{1}{c|}{N@10} & \multicolumn{1}{c}{R@1} & \multicolumn{1}{c}{R@5} & \multicolumn{1}{c|}{R@10} & \multicolumn{1}{c}{N@1} & \multicolumn{1}{c}{N@5} & \multicolumn{1}{c}{N@10} \\ \hline
\textsc{Mc}              & 0.189 & 0.364 & 0.407 & 0.189 & 0.284 & 0.298 & 0.170 & 0.313 & 0.347 & 0.170 & 0.247 & 0.258 \\
\textsc{Strnn}           & 0.162 & 0.255 & 0.287 & 0.162 & 0.213 & 0.223 & 0.123 & 0.209 & 0.246 & 0.123 & 0.169 & 0.180 \\
\textsc{DeepMove}        & 0.243 & 0.387 & 0.413 & 0.243 & 0.322 & 0.331 & 0.166 & 0.268 & 0.307 & 0.166 & 0.221 & 0.233 \\
\textsc{Lstpm}           & 0.235 & 0.436 & 0.492 & 0.235 & 0.342 & 0.361 & 0.205 & 0.366 & 0.416 & 0.205 & 0.292 & 0.309 \\
\textsc{Flashback}       & 0.219 & 0.368 & 0.423 & 0.219 & 0.299 & 0.317 & 0.209 & 0.387 & 0.447 & 0.209 & 0.305 & 0.325 \\
\textsc{PG\textsuperscript{2}Net}         
                        & 0.206 & 0.400 & 0.430 & 0.206 & 0.313 & 0.323 & 0.197 & 0.333 & 0.376 & 0.197 & 0.270 & 0.284 \\
\textsc{Plspl}          & 0.187 & 0.315 & 0.365 & 0.187 & 0.258 & 0.274 & 0.166 & 0.272 & 0.315 & 0.166 & 0.222 & 0.236 \\
\textsc{Gcdan}          & 0.188 & 0.311 & 0.344 & 0.188 & 0.256 & 0.267 & 0.171 & 0.297 & 0.343 & 0.171 & 0.239 & 0.253 \\
\textsc{Cslsl}          & 0.231 & 0.387 & 0.421 & 0.231 & 0.317 & 0.328 & 0.210 & 0.367 & 0.417 & 0.210 & 0.294 & 0.310  \\ 
\textsc{G-Flashback}& 0.219 & 0.371 & 0.428 & 0.219 & 0.300 & 0.319 & 0.209 & 0.387 & 0.441 & 0.209 & 0.304 & 0.322  \\
\textsc{Hmt-Grn} & 0.242 & 0.406 & 0.457 & 0.242 & 0.333 & 0.349 & 0.209 & 0.371 & 0.425 & 0.209 & 0.295 & 0.312  \\ 
\textsc{Fpgt} & 0.231 & 0.406 & 0.446 & 0.231 & 0.326 & 0.339 & 0.207 & 0.365 & 0.420 & 0.207 & 0.291 & 0.309  \\ \hline
\textsc{Hgarn}          & \textbf{0.273} & \textbf{0.520} & \textbf{0.575} & \textbf{0.273} & \textbf{0.405} & \textbf{0.423} & \textbf{0.234}          & \textbf{0.461} & \textbf{0.526} & \textbf{0.234} & \textbf{0.355} & \textbf{0.376} \\ \bottomrule
\end{tabular}
\end{threeparttable}       
}   
\label{table:mainresults}
\end{table*}

\renewcommand{\arraystretch}{1.2}
\begin{table*}[h] 
\centering
\caption{Comparison of different model sizes.}
\resizebox{.95\textwidth}{!}{
\begin{tabular}{ccccccccccc|c}
\toprule
\textsc{Strnn} & \textsc{DeepMove} & \textsc{Lstpm} & \textsc{Flashback} & \textsc{PG\textsuperscript{2}Net} & \textsc{Plspl} & \textsc{Gcdan} & \textsc{Cslsl} & \textsc{G-Flashback} & \textsc{Hmt-Grn} & \textsc{Fpgt} & \textsc{Hgarn} \\ \hline
75K   & 4.5M     & 13M   & 1.5M      & 11.8M & 15.8M & 22.3M & 16M   & 1.5M            & 50.3M  & 2.8M   & 13.3M \\ \bottomrule
\end{tabular}
}
\label{table:modelsizes}
\end{table*}

\subsection{Baselines \& Experimental Details}

\begin{itemize}[leftmargin=*]
    \item \textbf{\textsc{Mc}} \cite{rendle2010factorizing} is a widely used sequential prediction approach which models transition patterns based on visited locations. 
    \item \textbf{\textsc{Strnn}} \cite{liu2016predicting} is an \textsc{Rnn}-based model that incorporates the spatial-temporal contexts by leveraging transition matrices.
    \item \textbf{\textsc{DeepMove}} \cite{feng2018deepmove} uses attention mechanisms and an \textsc{Rnn} module to capture human mobility patterns.
    \item \textbf{\textsc{Lstpm}} \cite{sun2020go} introduces a non-local network and a geo-dilated \textsc{Lstm} to model human mobility patterns.
    \item \textbf{\textsc{Flashback}} \cite{yang2020location} is an \textsc{Rnn}-based model that leverages spatial and temporal intervals to compute an aggregated hidden state for prediction.
    \item \textbf{\textsc{Plspl}} \cite{wu2020personalized} incorporates activity information to learn user preferences and utilizes two \textsc{Lstm}s to capture human mobility patterns.
    \item \textbf{\textsc{PG\textsuperscript{2}Net}} \cite{li2021pg} learns users' group and personalized preferences with spatial-temporal attention-based Bi-\textsc{Lstm}.
    \item \textbf{\textsc{Gcdan}} \cite{dang2022predicting} leverages graph convolution to learn spatial-temporal representations and use dual-attention to model the sequential dependencies.
    \item \textbf{\textsc{Cslsl}} \cite{huang2022human} employs multi-task learning to model decision logic and two \textsc{Rnn}s to capture human mobility patterns.
    \item \textbf{\textsc{Graph-Flashback}} \cite{rao2022graph} adds \textsc{Gcn} to \textsc{Flashback} to enrich learned transition graph representations constructed based on defined similarity functions over embeddings from  the existing Knowledge Graph Embedding method.
    \item \textbf{\textsc{Hmt-Grn}} \cite{lim2022hierarchical} partitions the spatial map and performs a Hierarchical Beam Search to reduce the search space.
    \item \textbf{\textsc{Fpgt}} \cite{he2023feature} uses geographical and popularity feature-based POI grouping, together with a transformer network, to make the next POI recommendation.
\end{itemize}

For model evaluation, we adopt two commonly used metrics in the literature, Rec@K (Recall) and NDCG@K (Normalized Discounted Cumulative Gain), which are defined as: 
\begin{equation}
Recall@K=\frac{1}{|U|}\sum_{u \in U}\frac{|l_{u}^{|R_{u}|+1} \cap \widehat{l_{u, K}^{|R_{u}|+1}}|}{|l_{u}^{|R_{u}|+1}|}
\end{equation}
\begin{equation}
NDCG@K=\frac{1}{|U|}\sum_{u \in U}\sum_{i=1}^{K}\frac{|l_{u}^{|R_{u}|+1} \cap \widehat{l_{u, i}^{|R_{u}|+1}}|}{\log(i+1)}
\end{equation}
where $\widehat{l^{\cdot}_{u, k}}$ indicates the top $k$ predicted locations. 

We port all the baselines to our run time environment for fair comparisons based on their open-source codes. We carefully tuned their hyperparameters to get the best results. Additionally, unlike previous works that only evaluate overall model performances (main setting), we also conduct experiments under the \textit{recurring} and \textit{explorative} settings for more comprehensive performance evaluation. For the main and \textit{recurring} settings, we choose $K=\{1, 5, 10\}$ for evaluation. As the performance is generally poorer under the \textit{explorative} setting, we set $K=\{10, 20\}$. 

For the choice of hyperparameters, we set both $\lambda_{L}$ and $\lambda_{C}$ to 1, $\lambda_{r}$ to 0.6 for both datasets.
For embedding dimensions, we set $d=200$, $d^{u}=10$, $d^{t}=30$, $d^{g}=50$ and the dimension of encoders' hidden states are set to 600. Detailed reproducibility information can be found in Appendix~\ref{sec:reproduce}.

\subsection{Main Results}

Table \ref{table:mainresults} shows the performance comparison between different methods for next location prediction. \textsc{Hgarn} achieves state-of-the-art performance on both datasets across all metrics. 
Specifically, \textsc{Hgarn} outperforms the best baseline approach by 12-19\% on Recall@$K$  and NDCG@$K$ for NYC and 11-20\% for TKY. Its advantages become more significant as $K$ increases, validating the effectiveness of the hierarchical graph modeling and \textsc{MaHec} label for the next location prediction task. We also provide model sizes ({\em i.e.}, number of model’s trainable parameters) in Table \ref{table:modelsizes}. Since model sizes are data-specific, we use the NYC dataset for demonstration.

\renewcommand{\arraystretch}{1.2}
\begin{table}[h]
\caption{Performance under the \textit{recurring} setting.} 
\centering 

\resizebox{0.49\textwidth}{!}{
\Large
\begin{tabular}{c|cccccc|cccccc}
\toprule
\multirow{2}{*}{Recurring} 
& \multicolumn{6}{c|}{NYC} & \multicolumn{6}{c}{TKY} \\ \cline{2-13}
& R@1 & R@5 & R@10 & N@1 & N@5 & N@10 & R@1 & R@5 & R@10 & N@1 & N@5 & N@10 \\ \hline
\textsc{Mc}                         & 0.237 & 0.430 & 0.474 & 0.237 & 0.342 & 0.357 & 0.199 & 0.371 & 0.408 & 0.199 & 0.292 & 0.304  \\ 
\textsc{Strnn}                      & 0.189 & 0.248 & 0.259 & 0.189 & 0.248 & 0.259 & 0.162 & 0.273 & 0.316 & 0.162 & 0.221 & 0.235  \\ 
\textsc{DeepMove}                   & 0.243 & 0.387 & 0.413 & 0.243 & 0.322 & 0.331 & 0.209 & 0.332 & 0.372 & 0.209 & 0.275 & 0.288  \\ 
\textsc{Lstpm}                      & 0.282 & 0.513 & 0.533 & 0.282 & 0.409 & 0.428 & 0.249 & 0.433 & 0.484 & 0.249 & 0.348 & 0.364  \\ 
\textsc{Flashback}                  & 0.283 & 0.507 & 0.554 & 0.283 & 0.406 & 0.422 & 0.250 & 0.462 & 0.527 & 0.250 & 0.363 & 0.384  \\ 
\textsc{PG\textsuperscript{2}Net}   & 0.285 & 0.492 & 0.526 & 0.285 & 0.398 & 0.409 & 0.252 & 0.411 & 0.459 & 0.252 & 0.338 & 0.354  \\ 
\textsc{Plspl}                      & 0.251 & 0.413 & 0.450 & 0.251 & 0.340 & 0.352 & 0.209 & 0.336 & 0.384 & 0.209 & 0.277 & 0.292  \\ 
\textsc{Gcdan}                      & 0.242 & 0.405 & 0.439 & 0.242 & 0.331 & 0.342 & 0.227 & 0.389 & 0.436 & 0.227 & 0.315 & 0.330 \\ 
\textsc{Cslsl}                      & 0.288 & 0.498 & 0.542 & 0.288 & 0.404 & 0.418 & 0.254 & 0.457 & 0.511 & 0.254 & 0.364 & 0.382  \\ 
\textsc{G-Flashback}                & 0.282 & 0.509 & 0.562 & 0.282 & 0.406 & 0.423 & 0.252 & 0.463 & 0.527 & 0.252 & 0.364 & 0.385  \\
\textsc{Hmt-Grn}                    & 0.299 & 0.514 & 0.553 & 0.299 & 0.417 & 0.430 & 0.245 & 0.446 & 0.508 & 0.245 & 0.352 & 0.372  \\ 
\textsc{Fpgt}                 & 0.293 & 0.491 & 0.531 & 0.293 & 0.401 & 0.414 & 0.260 & 0.435 & 0.497 & 0.260 & 0.354 & 0.374  \\ \hline
\textsc{Hgarn}                & \textbf{0.319} & \textbf{0.633} & \textbf{0.713} & \textbf{0.319} & \textbf{0.487} & \textbf{0.514} & \textbf{0.278} & \textbf{0.552} & \textbf{0.631} & \textbf{0.278} & \textbf{0.424} & \textbf{0.450}  \\ \bottomrule
\end{tabular}          
}
\label{table:recurringresults}
\end{table}

In addition, we also evaluate different models separately in the \textit{recurring} and \textit{explorative} settings, with results shown in Tables~\ref{table:recurringresults} and \ref{table:explorativeresults}\footnote{The \textsc{Mc}'s results are all zeros and thus deleted from the table.}, respectively. In the \textit{recurring} setting, \textsc{Hgarn} outperforms all baselines significantly. In the \textit{explorative} setting, the overall prediction performance is much lower than those in the main and \textit{recurring} settings, which is intuitive because of the inherent difficulty of predicting unseen locations. A possible approach to improving the prediction in the \textit{explorative} setting is to model the dependencies between locations. 
In addition, due to the larger number of locations in the TKY dataset, the hierarchical graph modeling may introduce noise, making our model less effective in ranking the predicted locations. The above hypotheses may also be why our model performs better in Recall but are not consistently better in NDCG. 

\renewcommand{\arraystretch}{1.2}
\begin{table}[h]
\caption{Performance under the \textit{explorative} setting.} 
\centering 

\resizebox{0.47\textwidth}{!}{
\Large
\begin{tabular}{c|cccc|cccc}
\toprule
\multirow{2}{*}{Explorative} 
& \multicolumn{4}{c|}{NYC} & \multicolumn{4}{c}{TKY} \\ \cline{2-9}
& R@10 & R@20 & N@10 & N@20 & R@10 & R@20 & N@10 & N@20 \\ \hline
\textsc{Strnn}                          & 0.066 & 0.071 & 0.031 & 0.033 & 0.047 & 0.064  & 0.021 & 0.026  \\ 
\textsc{DeepMove}                       & 0.064 & 0.112 & 0.036 & 0.049  & 0.04 & 0.051  & 0.020 & 0.031 \\ 
\textsc{Lstpm}                          & 0.091 & 0.115 & 0.052 & 0.058 & 0.067 & 0.090 & 0.041 & 0.047 \\ 
\textsc{Flashback}                      & 0.083 & 0.109 & 0.045 & 0.051 & 0.053 & 0.072 & 0.028 & 0.032  \\ 
\textsc{PG\textsuperscript{2}Net}       & 0.046 & 0.054 & 0.021 & 0.023 & 0.056 & 0.065 & 0.029 & 0.032  \\ 
\textsc{Plspl}                          & 0.051 & 0.061 & 0.029 & 0.032 & 0.056 & 0.065 & 0.026 & 0.032  \\ 
\textsc{Gcdan}                          & 0.049 & 0.056 & 0.025 & 0.027 & 0.036 & 0.048 & 0.020 & 0.023  \\ 
\textsc{Cslsl}                          & 0.078 & 0.115 & 0.048 & 0.057 & 0.062 & 0.095 & 0.030 & 0.038  \\ 
\textsc{Graph-Flashback}                & 0.078 & 0.104 & 0.044 & 0.051 & 0.053 & 0.072 & 0.028 & 0.032  \\ 
\textsc{Hmt-Grn}                        & 0.081 & 0.102 & 0.052 & 0.058 & 0.072 & 0.098 & \textbf{0.050} & \textbf{0.057} \\ 
\textsc{Fpgt}                     & 0.096 & 0.112 & 0.046 & 0.050 & 0.076 & 0.102 & 0.040 & 0.046 \\ \hline
\textsc{Hgarn}                     & \textbf{0.102} & \textbf{0.135} & \textbf{0.054} & \textbf{0.062} & \textbf{0.081} & \textbf{0.120} & 0.037 & 0.047  \\ \bottomrule
\end{tabular}          
}
\label{table:explorativeresults}
\end{table}

\subsection{Ablation Study} \label{res:ablation}

\begin{figure}[h]
    \centering
    \includegraphics[width=0.49\textwidth]{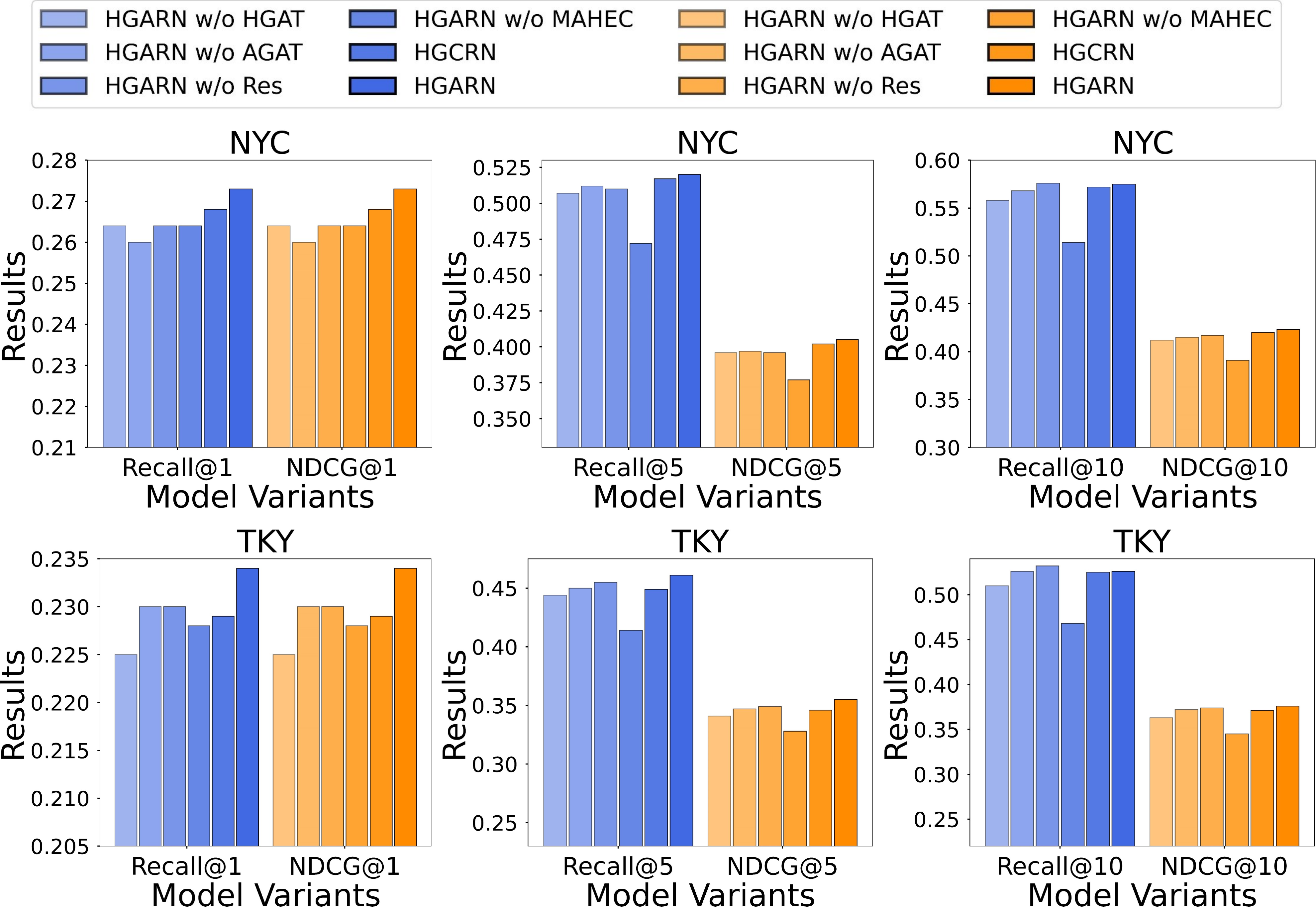}
    \caption{Ablation study results comparison.}
    \label{fig:ablation}
\end{figure}

To investigate the effectiveness of each component of \textsc{Hgarn}, we conduct an ablation study considering the following 6 variants of \textsc{Hgarn}:
\begin{itemize}[leftmargin=*]
\item \textsc{Hgarn} w/o \textsc{HGat} is the variant that contains only the \textit{temporal module} for next location prediction;
\item \textsc{Hgarn} w/o \textsc{AGat} is the variant whose \textit{hierarchical graph attention module} contains only the location layer and corresponding \textsc{Gat}s;
\item \textsc{Hgarn} w/o \textsc{Lal} is the variant whose \textit{hierarchical graph attention module} contains only the location layer and the activity layer;
\item \textsc{Hgarn} w/o \textsc{Res} is the variant that removes the residual connection in the \textit{temporal module};
\item \textsc{Hgarn} w/o \textsc{MaHec} is the variant that leverages original labels to optimize our model;
\item \textsc{Hgcrn} is the variant replacing the hierarchical graph attention with hierarchical graph convolution
\end{itemize}

The ablation study results are shown in Figure \ref{fig:ablation}. 
It is found that all performance metrics improve as more components are included. The gradual improvement in prediction results from \textsc{Hgarn} w/o \textsc{HGat} to \textsc{Hgarn} w/o \textsc{AGat}, and then to \textsc{Hgarn} clearly demonstrates the effectiveness of \textsc{Gat} in each layer of the hierarchical graph. In addition, adopting the \textsc{MaHec} label leads to a significant model performance improvements when $K$ are large, verifying its effectiveness despite the simplicity.

The comparison between \textsc{Hgarn} and \textsc{Hgcrn} shows that \textsc{Gat} can outperform \textsc{Gcn} for graph-based learning tasks through its self-attention mechanism. This has been supported by previous research \cite{lv2021we} demonstrating that \textsc{Gat} could achieve competitive performance through proper hyperparameter tuning and configuration compared to \textsc{Gcn}. Additionally, it provides more robust inductive capabilities that can incorporate newly available (unseen) nodes without retraining. Finally, \textsc{Gat} is more adaptable and scalable than \textsc{Gcn} due to its dynamic scheme that can automatically learn the importance of each node from a graph structure. The hierarchical design can help to overcome the over-smoothing problem \cite{zhao2019pairnorm} of \textsc{Gnn}s, and the dependencies between location nodes sharing the same activity across regions can be modeled.
In contrast, the \textsc{Hgarn} variant without the \textsc{Lal} layer (\textsc{Hgarn} w/o \textsc{Lal}) suffers from "information overload" from the location layer, as it aggregates excessive information without proper filtering. This results in a reduced ability to effectively capture both activity-to-activity and fine-grained activity-to-location dependencies, as evidenced by the ablation study. 
The localized-activity layer (\textsc{Lal}) in \textsc{Hgarn} addresses this issue by focusing the model on relevant activity-location relationships while suppressing noise from overly dense connections in the location layer, leading to more effective learning. This approach aligns with techniques explored in other hierarchical GNN models, where multi-layer aggregation schemes help filter irrelevant information and better capture hierarchical relationships in complex systems \cite{wu2020learning,guo2021hierarchical}.

\subsection{Hyperparameter Sensitivity Analysis}

\begin{figure}[h]
    \centering
    \includegraphics[width=0.49\textwidth]{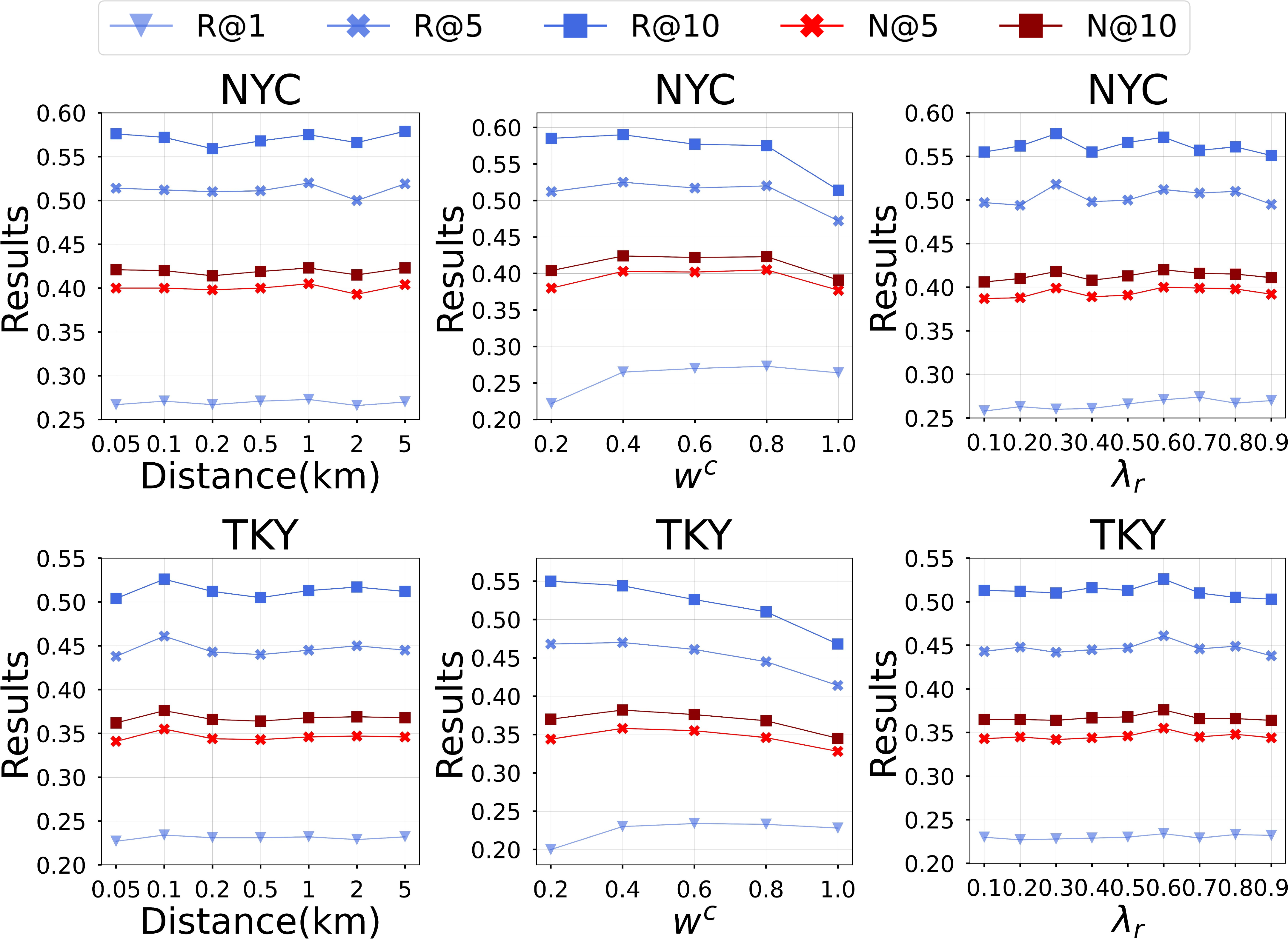}
    \caption{Sensitivity experiments results on two datasets.}
    \label{fig:sensitivity}
\end{figure}

We further study the sensitivity of a few key parameters by varying each parameter while keeping others constant. The results in Figure~\ref{fig:sensitivity} illustrate that the distance threshold $D^{h}$ affects the location dependencies; if the threshold is too high or too low, the prediction performance would be negatively affected. The best results are obtained at $D^{h}=1$km for NYC and $D^{h}=0.1$km for TKY, probably due to the higher density in TKY, as shown in Figure~\ref{fig:spatial_appendix}. The \textsc{MaHec} hyperparameter $w^{c}$ affects how the model treats the locations a user visited in the past. 
For both datasets, the results show an upward and then downward trend as $w^{c}$ increases, suggesting that a moderate amount of attention to the previously visited locations can produce the best model performance. This indicates that striking a balance between exploration and recurrence leads to optimal performance in overall mobility modeling.
$\lambda_{r}$ from Eq.~\eqref{eq16} affects how much activity information is fused in predicting the next location. Intuitively, a large value of $\lambda_{r}$ would introduce more noise, and a small value may result in ineffective utilization of activity information. The results confirm that the model achieves optimal performance with $\lambda_{r}$ at 0.6 for both the NYC and TKY datasets.
To summarize, the proposed \textsc{Hgarn} model demonstrates robustness across a range of parameter settings, with only small oscillations in performance as parameters vary. The results also demonstrate the model’s capacity to balance exploration and recurrence effectively, while integrating activity information, resulting in consistently strong performance across both datasets.

\subsection{Interpretability Analysis}

\subsubsection{How does the \textsc{MaHec} label work?}

\begin{figure}[h]
    \centering
    \includegraphics[width=0.47\textwidth]{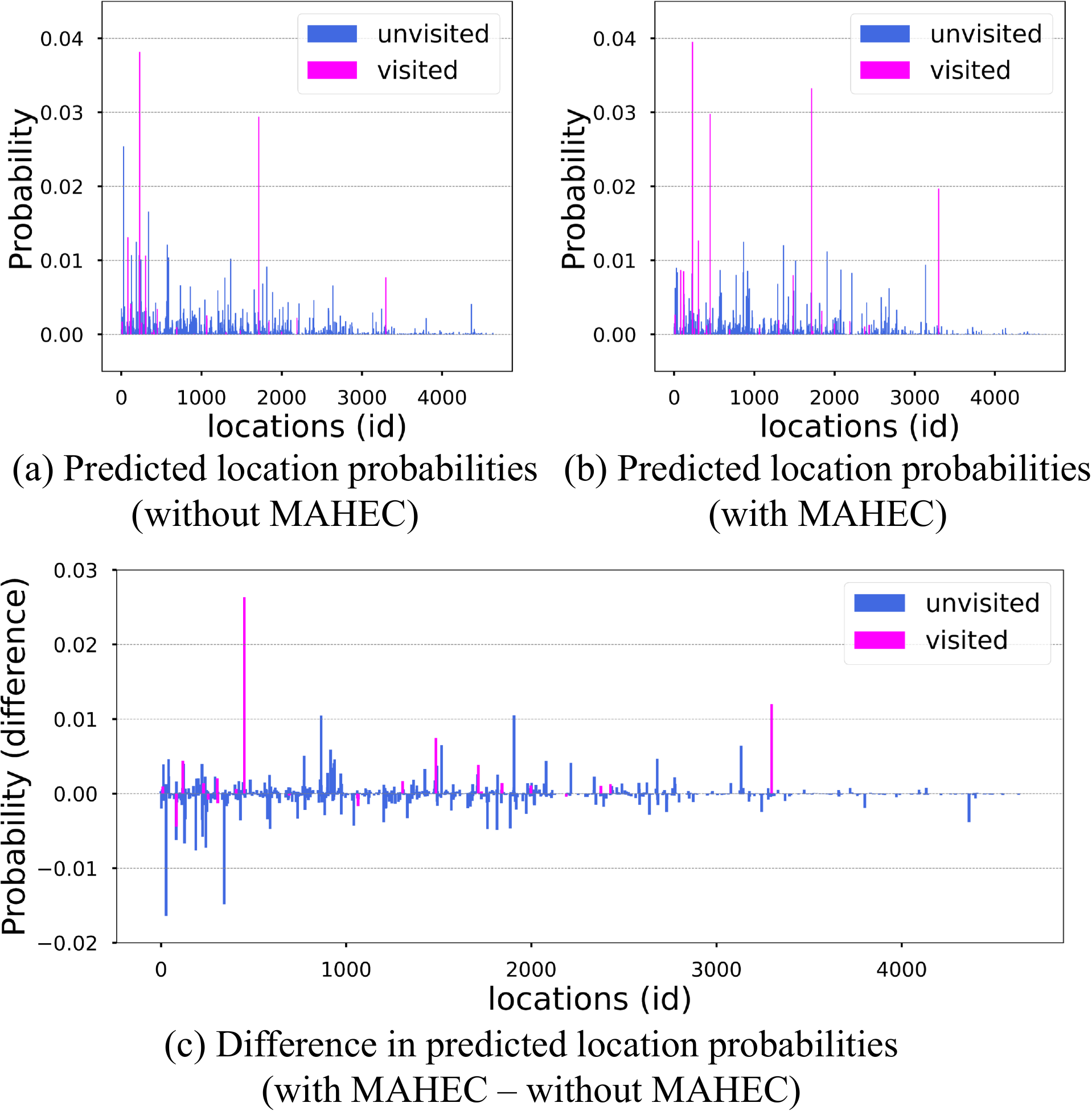}
    \caption{Comparison of predicted location probabilities for a user from NYC (with \textsc{MaHec} vs without \textsc{MaHec}).}
    \label{fig:mahec}
\end{figure}

To understand the mechanism of the \textsc{MaHec} label, we select an example user trajectory from the NYC dataset, visualize the predicted location probabilities based on ``\textsc{Hgarn},'' and ``\textsc{Hgarn} w/o \textsc{MaHec},'' and compare their differences. Figure \ref{fig:mahec} demonstrates the changes in predicted probabilities, where purple bars represent the locations in the user's observed trajectory (i.e., \textit{visited}) and blue bars indicate \textit{unvisited} locations. 
The results reveal that the use of \textsc{MaHec} labels increases the model's predicted probability for the next location across previously visited locations ({\em i.e.}, the probability difference for visited locations remains positive). 
This indicates that \textsc{MaHec} labels effectively guide the model to consider the user's past movements when predicting the next location through the model learning process.
The effectiveness of \textsc{MaHec} labels can also partially explain why the prediction performance of \textsc{Hgarn} significantly exceeds that of the baseline methods, especially under the \textit{recurring} setting.

\subsubsection{What the Hierarchical Graph learned?}

\begin{figure}[h]
    \centering
    \includegraphics[width=0.48\textwidth]{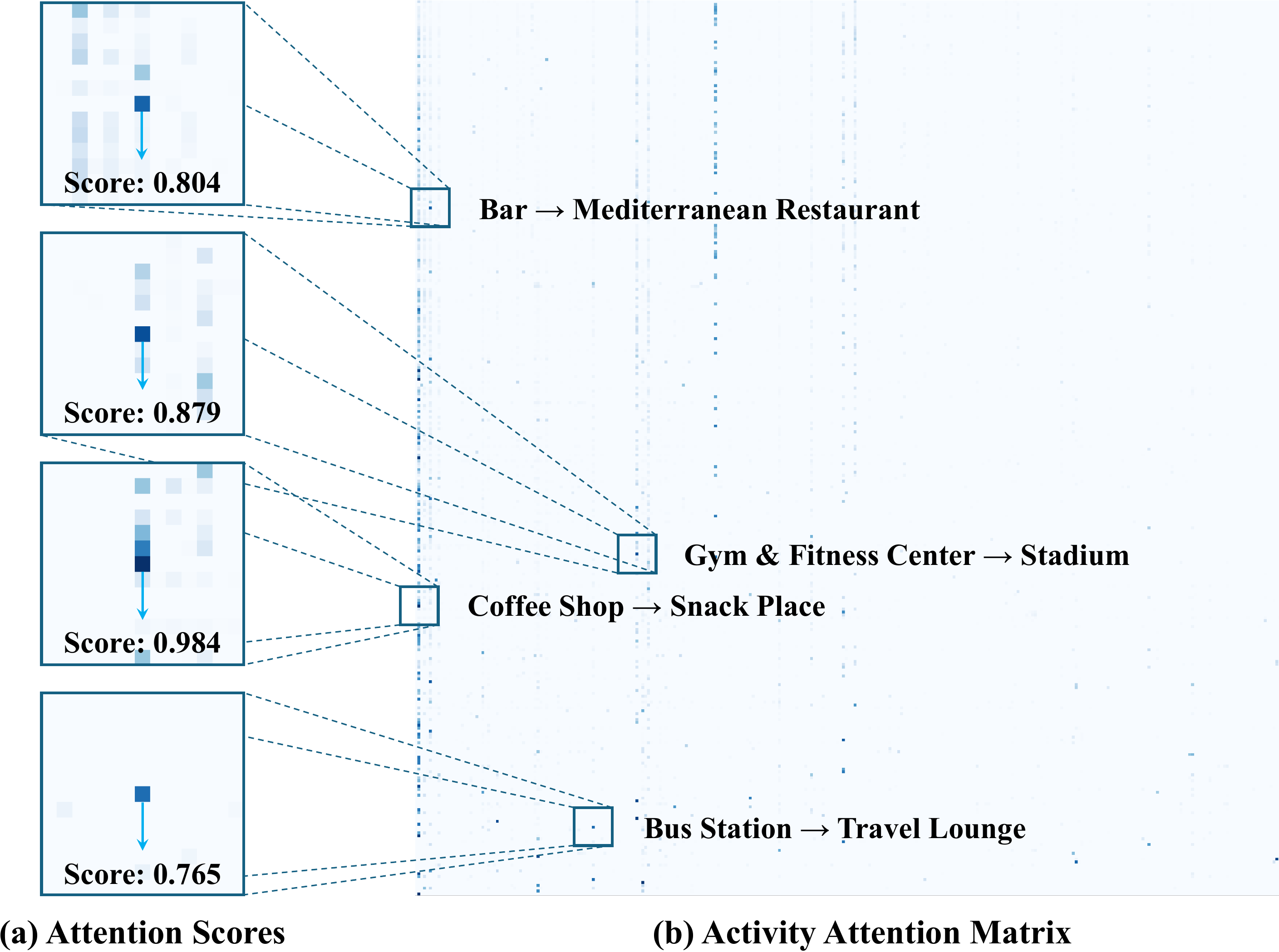}
    \caption{A visualization of activities' attentions and examples.}
    \label{fig:c_attn}
\end{figure}

Unlike other deep learning methods that may suffer from limited interpretability, \textsc{Hgarn} can be used to reveal the dependencies between activities through the learned Hierarchical Graph. We visualize one attention head of $\textsc{Gat}_{C}$'s sliced attention matrix to analyze the learned activity-activity dependencies. In Figure \ref{fig:c_attn}, we select four activity pairs to show the related activities and their corresponding attention scores. These activity pairs are consistent with common sense, such as the high dependencies between Gyms and Stadiums, or Bus Stops and Travel Lounges. These results have important implications for understanding human activity patterns and predicting future mobility behavior.

\section{Conclusion}

Both travel behavior theories and empirical evidence suggest that human mobility patterns largely depend on the need to participate in activities at different times of the day. Therefore, it is crucial to consider the latter when modeling the former. In this paper, we propose a \textit{Hierarchical Graph Attention Recurrent Network} (\textsc{Hgarn}) for activity-aware human mobility prediction.
Specifically, \textsc{Hgarn} introduces hierarchical graph attention mechanisms to model time-activity-location dependencies, and uses next activity prediction as an auxiliary task to further improve the main task of next location prediction. Furthermore, we propose a simple yet effective \textsc{MaHec} label that can guide our model to flexibly weigh the importance of a user's previously visited locations when predicting their future locations. Finally, based on two real-world LBSN datasets, we perform comprehensive experiments to demonstrate the superiority of \textsc{Hgarn}, considering both the \textit{recurring} and \textit{explorative} settings. We find that introducing activity information can effectively improve the model's prediction performance, and the learned attention weights can reveal meaningful behavioral insights.

Future work should prioritize improving human mobility prediction in \textit{explorative} settings, where users visit new, unvisited POIs. Developing models that can better infer these complex time-activity-location relationships, even without prior visitation history, will be crucial. Another key challenge is the cold-start problem, where the model must handle new users, locations, or activities introduced into the system. Addressing this could involve leveraging shared features like activity or contextual embeddings for initializing new entities, minimizing retraining needs. From a system design perspective, future models should aim for a lightweight and modular structure. These advancements, combined with a deeper understanding of human decision-making in travel behavior, will not only improve the model's interpretability but also contribute to more intelligent, user-centered transportation systems that offer personalized and efficient travel recommendations.

\section*{Acknowledgments}
This research is supported by the National Natural Science Foundation of China (NSFC 42201502) and Seed Funding for Strategic Interdisciplinary Research Scheme at the University of Hong Kong (102010057).

\section{Appendix}

\setcounter{table}{0}
\renewcommand{\thetable}{A\arabic{table}}
\setcounter{figure}{0}
\renewcommand{\thefigure}{A\arabic{figure}}

\subsection{Data Preprocessing}

\renewcommand{\arraystretch}{1.2}
\begin{table}[h]
\caption{Check-in data format.} 
\centering 

\resizebox{0.475\textwidth}{!}{
\begin{tabular}{c|c} \toprule
  \textbf{ID}    & \textbf{Data (e.g.)} \\
\hline
  User ID                         & 470 \\
  Venue ID                        & 49bbd6c0f964a520f4531fe3 \\
  Venue category ID               & 4bf58dd8d48988d127951735 \\
  Venue category name             & Arts \& Crafts Store \\
  Latitude                        & 40.719810375488535 \\
  Longitude                       & -74.00258103213994 \\
  Timezone offset in minutes      & -240 \\
  UTC time                        & Tue Apr 03 18:00:09 +0000 2012 \\
\bottomrule
\end{tabular}          
}
\label{table:datasets}
\end{table}

The adopted dataset contains check-in data in New York city (NYC) and Tokyo (TKY) collected from Foursquare from 12 April 2012 to 16 February 2013. The data format is shown in Table~\ref{table:datasets}. NYC contains 227428 check-ins in New York city, TKY contains 573703 check-ins in Tokyo.

\begin{figure}[h]
    \centering
    \includegraphics[width=.46\textwidth]{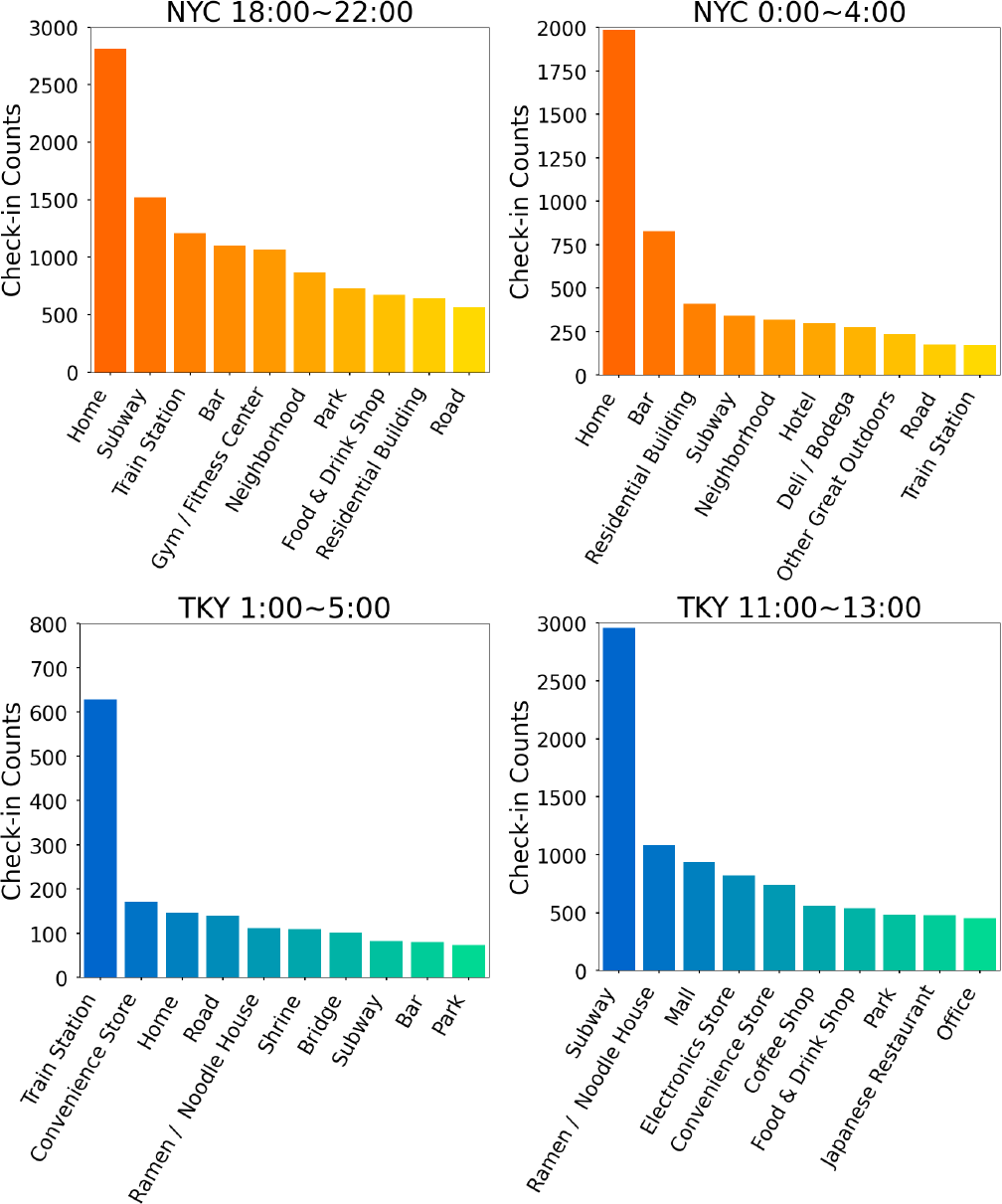}
    \caption{Activity distribution statistics in New York and Tokyo at two different time periods.}
    \label{fig:timeact}
\end{figure}

Figure \ref{fig:timeact} shows the frequency distribution across different activity types during two selected time periods in NYC and TKY. We removed TKY's top-1 activity Train Station (with 12468 check-ins) at 11:00~13:00 for better visualization. The results demonstrate a strong temporal dependency between activities.

\begin{figure}[h]
    \centering
    \includegraphics[width=.4\textwidth]{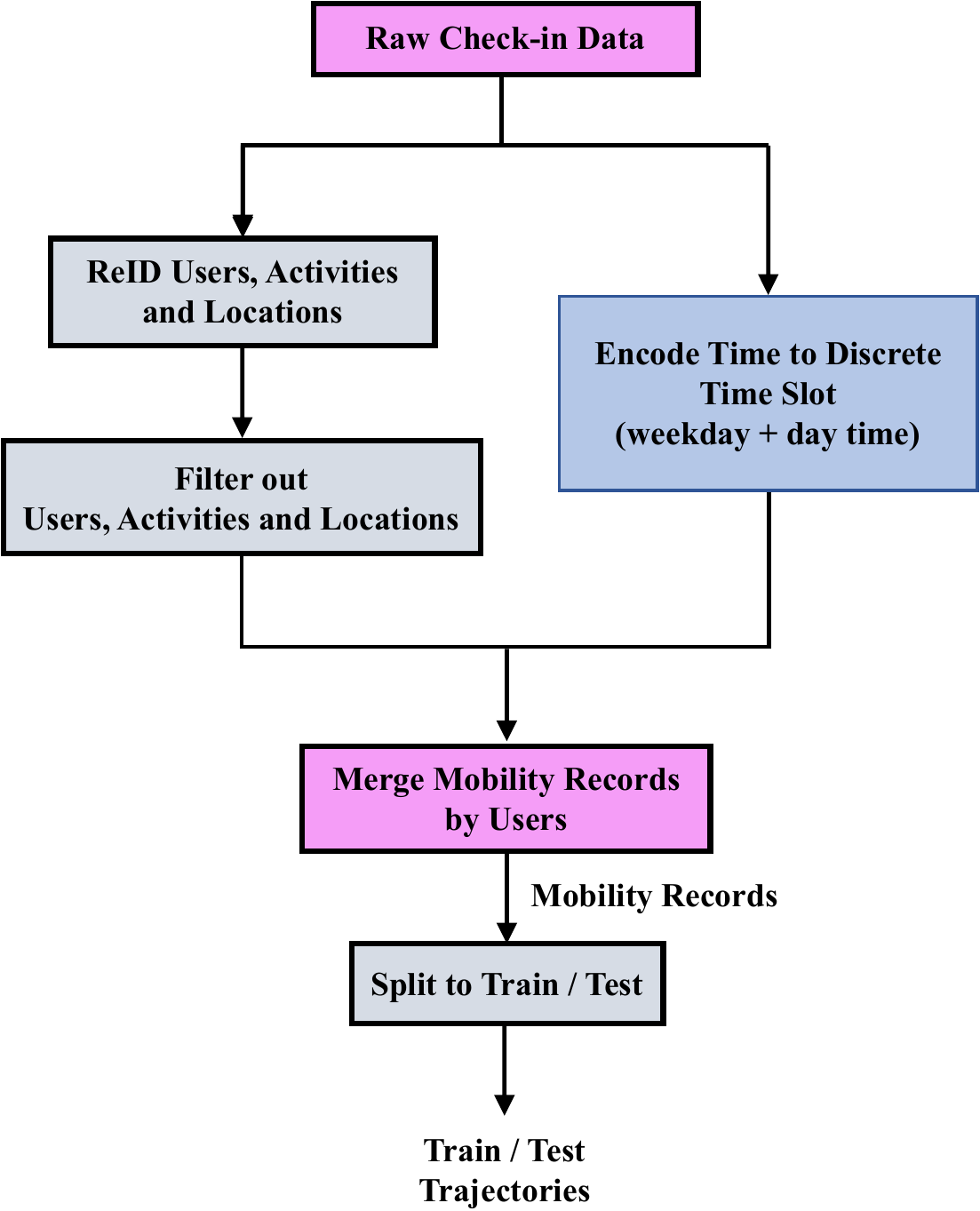}
    \caption{Data preprocessing workflow.}
    \label{fig:datapre}
\end{figure}

The data preprocessing flow is illustrated in Figure \ref{fig:datapre}. Based on the raw check-in data, we identify users, locations, and activities and filter out elements with fewer than 10 records. Next, we convert the continuous time to discrete time intervals, and merge mobility records by users to form trajectories. Lastly, we split all trajectories into training and testing data for model fitting and evaluation.

\subsection{Reproducibility} \label{sec:reproduce}

For reproducibility of our study, we provide the specific information about computing devices and detailed hyperparameter settings used in our experiments. The source codes of this study are available at \url{https://github.com/YihongT/HGARN}.

All models (including \textsc{Hgarn} and other baselines) with learnable parameters are trained on a desktop with Intel(R) Xeon(R) Platinum 8375C CPU @2.90GHz $\times$ 64, 125Gi RAM, NVIDIA GeForce RTX 3090 $\times$ 8, 4TB SSD. We implement \textsc{Hgarn} based on Pytorch. Parameters of \textsc{Hgarn} are randomly initialized and optimized using the Adam optimizer with a learning rate of 2e-4, decaying by 0.8 with each epoch. \textsc{Hgarn} is trained in 80 epochs.

For hyperparameter settings, we set $\lambda_{L}$=$\lambda_{C}$=1, and dimensions $d^{g}=50$, $d=200$, $d^{u}=20$, $d^{t}=30$. Location and activity encoders have hidden states with dimensions of 600. The above settings remain the same for all experiments. For the main and \textit{recurring} settings, we employ 2 attention heads and set the dropout to 0.1. For the \textit{explorative} setting, to prevent overfitting, we set the number of attention heads to 1 and the dropout to 0.6. For the NYC's main and \textit{recurring} settings, we set $D^{h}$ to 1, $w^{c}$ to 0.8, and $\lambda_{r}$ to 0.6; for the \textit{explorative} setting, they are set to 0.1, 0.9, and 1, respectively. For the TKY's main and \textit{recurring} settings, $D^{h}$, $w^{c}$, and $\lambda_{r}$ are set to 0.1, 0.6 and 0.6, respectively; for the \textit{explorative} setting, they are set to 0.1, 1, 0.5, respectively.

\subsection{An Example of User Embedding}
\label{appx:embedding}

\begin{figure}[h]
    \centering
    \includegraphics[width=.48\textwidth]{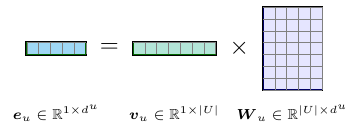}
    \caption{An illustration of user embedding.}
    \label{fig:onehot_embedding}
\end{figure}

In Figure \ref{fig:onehot_embedding}, we demonstrate a user $u$’s embedding process, where $\boldsymbol{e}_u$ is the learned user embedding vector. Similar processes apply to the location $l$, activity $c$, and time slot $t$ to obtain corresponding embedding vectors $\boldsymbol{e}_l$, $\boldsymbol{e}_c$, and $\boldsymbol{e}_t$, respectively.

\subsection{Training Algorithm}

The training process of \textsc{Hgarn} is detailed in Algorithm~\ref{alg:1}.

\begin{algorithm}
  \caption{Training algorithm of \textsc{Hgarn}}
  \begin{algorithmic}[1]\label{alg:1}
    \REQUIRE Observed trajectories $R$, the corresponding sets of users $U$, locations, $L$, activities $C$ and time intervals $T$. 

    \STATE Initialize \textsc{Hgarn}'s parameters and set hyperparameters
    \STATE /* \textbf{Hierarchical Graph Construction} */
    \STATE Construct the hierarchical graph $\mathcal{G}=(\mathcal{V},\mathcal{E})$ using Eqs.~\eqref{eq3}--\eqref{eq6}, where $\mathcal{V}=\mathcal{V}^{L} \cup \mathcal{V}^{C} \cup \mathcal{V}^{C^\prime}$ and $\mathcal{E}=\{A^{L}$,$A^{C}$,$A^{LC^\prime}$,$A^{CC^\prime}\}$
    \WHILE{not converge} 
        \FOR{batch}
            \STATE /* \textbf{Embedding Module} */
            \STATE Compute the embeddings for users $\boldsymbol{e}_{U}$, locations $\boldsymbol{e}_{L}$, activities $\boldsymbol{e}_{C}$, and time intervals $\boldsymbol{e}_{T}$ using Eq.~\eqref{eq2}
            \STATE /* \textbf{Hierarchical Graph Attention Module} */
            \STATE Compute $\boldsymbol{H}^{L}$, $\boldsymbol{H}^{C^\prime}$ using Eqs.~\eqref{eq7}--\eqref{eq10}, and remove the appropriate rows as described in the main text to obtain $\boldsymbol{H}^{C}$.
            \STATE /* \textbf{Temporal Module} */
            \STATE Calculate the input $\boldsymbol{X}^{C,i}_{U}$ and $\boldsymbol{X}^{L,i}_{U}$ for the activity and location encoders using Eqs.~\eqref{eq11}--\eqref{eq12}
            \STATE Recurrently encode the input $\boldsymbol{X}^{C,i}_{U}$ and $\boldsymbol{X}^{L,i}_{U}$ using Eqs.~\eqref{eq13}--\eqref{eq14} to obtain the final hidden states of activity and location encoder as $\boldsymbol{h}^{C}_{U}$, $\boldsymbol{h}^{L}_{U}$
            \STATE Obtain the predicted probability distribution by applying a $\operatorname{Softmax}$ function over the outputs of Eqs.~\eqref{eq15}--\eqref{eq16}
            \STATE /* \textbf{Construct \textsc{MaHec} Labels} */
            \STATE Compute \textsc{MaHec} labels $\textsc{MaHec}_{C}$ and $\textsc{MaHec}_{L}$ based on $R_{U}^{L}$, $R_{U}^{C}$ using Eqs.~\eqref{eq17}--\eqref{eq18}
            \STATE Compute the total prediction loss $\mathcal{L}$ as a combination of location loss $\mathcal{L}_L$ and activity loss $\mathcal{L}_C$ using Eqs.~\eqref{eq19}--\eqref{eq20}
        \ENDFOR
        \STATE Perform gradient descent to update model parameters
    \ENDWHILE
  \end{algorithmic}
\end{algorithm}

\subsection{Full Numerical Results}

In this section, we show the complete numerical results for the figures presented in the main text. Table \ref{table:ablations} contains complete numerical results for our ablation study. Tables \ref{table:sensitivity_dist}, \ref{table:sensitivity_res} and \ref{table:sensitivity_w} shows the complete sensitivity experiment results for $D^{h}$, $\lambda_{r}$ and $w^{c}$, respectively.

\renewcommand{\arraystretch}{1.2}
\begin{table}[h]
\caption{Full results of the ablation study.} 
\centering 

\resizebox{0.485\textwidth}{!}{
\Large
\begin{tabular}{c|c|cccccc}
\toprule
\multicolumn{2}{c|}{Ablations} & \multicolumn{1}{c}{w/o \textsc{HGat}} & \multicolumn{1}{c}{w/o \textsc{AGat}} & \multicolumn{1}{c}{w/o \textsc{Res}} & \multicolumn{1}{c}{w/o \textsc{MaHec}} & \multicolumn{1}{c}{\textsc{Hgcrn}} & \multicolumn{1}{c}{\textsc{Hgarn}}\\ \hline
\multirow{6}{*}{\rotatebox{90}{NYC}} 
& R@1  & 0.264 & 0.260 & 0.264 & 0.264 & 0.268 & 0.273 \\ \cline{2-2} 
& R@5  & 0.507 & 0.512 & 0.510 & 0.472 & 0.517 & 0.520 \\ \cline{2-2} 
& R@10 & 0.558 & 0.568 & 0.576 & 0.514 & 0.572 & 0.575 \\ \cline{2-2} 
& N@1  & 0.264 & 0.260 & 0.264 & 0.264 & 0.268 & 0.273 \\ \cline{2-2} 
& N@5  & 0.396 & 0.397 & 0.396 & 0.377 & 0.402 & 0.405 \\ \cline{2-2} 
& N@10 & 0.412 & 0.415 & 0.417 & 0.391 & 0.420 & 0.423 \\ \cline{2-2} 
\hline
\multirow{6}{*}{\rotatebox{90}{TKY}} 
& R@1  & 0.225 & 0.230 & 0.230 & 0.228 & 0.229 & 0.234 \\ \cline{2-2} 
& R@5  & 0.444 & 0.450 & 0.455 & 0.414 & 0.449 & 0.461 \\ \cline{2-2} 
& R@10 & 0.510 & 0.526 & 0.532 & 0.468 & 0.525 & 0.526 \\ \cline{2-2} 
& N@1  & 0.225 & 0.230 & 0.230 & 0.228 & 0.229 & 0.234 \\ \cline{2-2} 
& N@5  & 0.341 & 0.347 & 0.349 & 0.328 & 0.346 & 0.355 \\ \cline{2-2} 
& N@10 & 0.363 & 0.372 & 0.374 & 0.345 & 0.371 & 0.376 \\ \cline{2-2} 
\bottomrule
\end{tabular}          
}
\label{table:ablations}
\end{table}

\renewcommand{\arraystretch}{1.2}
\begin{table}[h]
\caption{Full results of $w^{c}$'s sensitivity experiments.} 
\centering 

\resizebox{0.45\textwidth}{!}{
\small
\begin{tabular}{c|c|ccccc}
\toprule
\multicolumn{2}{c|}{$w^{c}$ (\textsc{MaHec})} & \multicolumn{1}{c}{\text{        }0.2\text{        }} & \multicolumn{1}{c}{\text{        }0.4\text{        }} & \multicolumn{1}{c}{\text{        }0.6\text{        }} & \multicolumn{1}{c}{\text{        }0.8\text{        }} & \multicolumn{1}{c}{\text{        }1.0\text{        }} \\ \hline
\multirow{6}{*}{\rotatebox{0}{NYC}} 
& R@1  & 0.222 & 0.265 & 0.270 & 0.273 & 0.264 \\ \cline{2-2} 
& R@5  & 0.512 & 0.525 & 0.517 & 0.520 & 0.472 \\ \cline{2-2} 
& R@10 & 0.585 & 0.590 & 0.577 & 0.575 & 0.514 \\ \cline{2-2} 
& N@1  & 0.222 & 0.265 & 0.270 & 0.273 & 0.264 \\ \cline{2-2} 
& N@5  & 0.380 & 0.403 & 0.402 & 0.405 & 0.377 \\ \cline{2-2} 
& N@10 & 0.404 & 0.424 & 0.422 & 0.423 & 0.391 \\ \cline{2-2} 
\hline
\multirow{6}{*}{\rotatebox{0}{TKY}} 
& R@1  & 0.200 & 0.230 & 0.234 & 0.233 & 0.228 \\ \cline{2-2} 
& R@5  & 0.468 & 0.470 & 0.461 & 0.445 & 0.414 \\ \cline{2-2} 
& R@10 & 0.550 & 0.544 & 0.526 & 0.510 & 0.468 \\ \cline{2-2} 
& N@1  & 0.200 & 0.230 & 0.234 & 0.233 & 0.228 \\ \cline{2-2} 
& N@5  & 0.344 & 0.358 & 0.355 & 0.346 & 0.328 \\ \cline{2-2} 
& N@10 & 0.370 & 0.382 & 0.376 & 0.368 & 0.345 \\ \cline{2-2} 
\bottomrule
\end{tabular}          
}
\label{table:sensitivity_w}
\end{table}

\renewcommand{\arraystretch}{1.2}
\begin{table}[h]
\caption{Full results of $D^{h}$'s sensitivity experiments.}
\centering 

\resizebox{0.485\textwidth}{!}{
\Large
\begin{tabular}{c|c|ccccccc}
\toprule
\multicolumn{2}{c|}{$D^{h}$} & \multicolumn{1}{c}{0.05} & \multicolumn{1}{c}{0.1} & \multicolumn{1}{c}{0.2} & \multicolumn{1}{c}{0.5} & \multicolumn{1}{c}{1} & \multicolumn{1}{c}{2} & \multicolumn{1}{c}{5}                                                             \\ \hline
\multirow{6}{*}{\rotatebox{90}{NYC}} 
& R@1  & 0.267 & 0.271 & 0.267 & 0.271 & 0.273 & 0.266 & 0.270 \\ \cline{2-2} 
& R@5  & 0.514 & 0.512 & 0.510 & 0.511 & 0.520 & 0.500 & 0.519 \\ \cline{2-2} 
& R@10 & 0.576 & 0.572 & 0.559 & 0.568 & 0.575 & 0.566 & 0.579 \\ \cline{2-2} 
& N@1  & 0.267 & 0.271 & 0.267 & 0.271 & 0.273 & 0.266 & 0.270 \\ \cline{2-2} 
& N@5  & 0.400 & 0.400 & 0.398 & 0.400 & 0.405 & 0.393 & 0.404  \\ \cline{2-2} 
& N@10 & 0.421 & 0.420 & 0.414 & 0.419 & 0.423 & 0.415 & 0.423 \\ \cline{2-2} 
\hline
\multirow{6}{*}{\rotatebox{90}{TKY}} 
& R@1  & 0.227 & 0.234 & 0.231 & 0.231 & 0.232 & 0.229 & 0.232 \\ \cline{2-2} 
& R@5  & 0.438 & 0.461 & 0.443 & 0.440 & 0.445 & 0.450 & 0.445 \\ \cline{2-2} 
& R@10 & 0.504 & 0.526 & 0.512 & 0.505 & 0.513 & 0.517 & 0.512 \\ \cline{2-2} 
& N@1  & 0.227 & 0.234 & 0.231 & 0.231 & 0.232 & 0.229 & 0.232 \\ \cline{2-2} 
& N@5  & 0.341 & 0.355 & 0.344 & 0.343 & 0.346 & 0.347 & 0.346 \\ \cline{2-2} 
& N@10 & 0.362 & 0.376 & 0.366 & 0.364 & 0.368 & 0.369 & 0.368 \\ \cline{2-2} 
\bottomrule
\end{tabular}          
}
\label{table:sensitivity_dist}
\end{table}

\renewcommand{\arraystretch}{1.4}
\begin{table}[h]
\caption{Full results of $\lambda_{r}$'s sensitivity experiments.} 
\centering 

\resizebox{0.485\textwidth}{!}{
\Large
\begin{tabular}{c|c|ccccccccc}
\toprule
\multicolumn{2}{c|}{$\lambda_{r}$}& \multicolumn{1}{c}{0.1} & \multicolumn{1}{c}{0.2} & \multicolumn{1}{c}{0.3} & \multicolumn{1}{c}{0.4} & \multicolumn{1}{c}{0.5} & \multicolumn{1}{c}{0.6} & \multicolumn{1}{c}{0.7} & \multicolumn{1}{c}{0.8} & \multicolumn{1}{c}{0.9} \\ \hline
\multirow{6}{*}{\rotatebox{90}{NYC}} 
& R@1  & 0.258 & 0.263 & 0.260 & 0.261 & 0.266 & 0.271 & 0.274 & 0.267 & 0.270 \\ \cline{2-2} 
& R@5  & 0.497 & 0.494 & 0.518 & 0.498 & 0.500 & 0.512 & 0.508 & 0.510 & 0.495 \\ \cline{2-2} 
& R@10 & 0.555 & 0.562 & 0.576 & 0.555 & 0.566 & 0.572 & 0.557 & 0.561 & 0.551 \\ \cline{2-2} 
& N@1  & 0.258 & 0.263 & 0.260 & 0.261 & 0.266 & 0.271 & 0.274 & 0.267 & 0.270 \\ \cline{2-2} 
& N@5  & 0.387 & 0.388 & 0.399 & 0.389 & 0.391 & 0.400 & 0.399 & 0.398 & 0.392 \\ \cline{2-2} 
& N@10 & 0.406 & 0.410 & 0.418 & 0.408 & 0.413 & 0.420 & 0.416 & 0.415 & 0.411 \\ \cline{2-2} 
\hline
\multirow{6}{*}{\rotatebox{90}{TKY}} 
& R@1  & 0.230 & 0.227 & 0.228 & 0.229 & 0.230 & 0.234 & 0.229 & 0.233 & 0.232 \\ \cline{2-2} 
& R@5  & 0.443 & 0.448 & 0.442 & 0.445 & 0.447 & 0.461 & 0.446 & 0.449 & 0.438 \\ \cline{2-2} 
& R@10 & 0.513 & 0.512 & 0.510 & 0.516 & 0.513 & 0.526 & 0.510 & 0.505 & 0.503 \\ \cline{2-2} 
& N@1  & 0.230 & 0.227 & 0.228 & 0.229 & 0.230 & 0.234 & 0.229 & 0.233 & 0.232 \\ \cline{2-2} 
& N@5  & 0.343 & 0.345 & 0.342 & 0.344 & 0.346 & 0.355 & 0.345 & 0.348 & 0.344 \\ \cline{2-2} 
& N@10 & 0.365 & 0.365 & 0.364 & 0.367 & 0.368 & 0.376 & 0.366 & 0.366 & 0.364 \\ \cline{2-2} 
\bottomrule
\end{tabular}          
}
\label{table:sensitivity_res}
\end{table}

\bibliographystyle{IEEEtran}
\bibliography{main}

\end{document}